\documentclass[runningheads]{llncs}
\usepackage{graphicx}
\usepackage[table]{xcolor}
\usepackage{tabularx}
\usepackage[pagebackref,breaklinks,colorlinks]{hyperref}
\usepackage{multirow,booktabs}
\hypersetup{citecolor=[RGB]{119,185,0}}
\usepackage{tikz}
\usepackage{comment}
\usepackage{amsmath,amssymb} 
\usepackage{pifont}
\usepackage{color}
\usepackage{xurl}

\usepackage[accsupp]{axessibility}  

\let\svthefootnote\thefootnote
\newcommand\blankfootnote[1]{%
  \let\thefootnote\relax\footnotetext{#1}%
  \let\thefootnote\svthefootnote%
}

\usepackage{duckuments}
\begin{document}
\pagestyle{headings}
\mainmatter
\def\ECCVSubNumber{1817}  

\title{PersFormer: 3D Lane Detection via Perspective Transformer and the OpenLane Benchmark} 

\titlerunning{PersFormer and OpenLane}
%
\author{Li Chen$^{1\ast\dagger}$ \and
Chonghao Sima$^{1\ast}$ \and
Yang Li$^{1\ast}$ \and
Zehan Zheng$^1$ \and
Jiajie Xu$^1$ \and\\
Xiangwei Geng$^1$ \and
Hongyang Li$^{1,2\dagger}$ \and
Conghui He$^1$ \and
Jianping Shi$^3$ \and
Yu Qiao$^1$ \and
Junchi Yan$^{1,2}$
}
%
\authorrunning{L. Chen et al.}
%
\institute{
{$^{1}$}~Shanghai AI Laboratory \quad
{$^{2}$}~Shanghai Jiao Tong University \\
{$^{3}$}~SenseTime Research \\
\email{\{lichen,simachonghao,liyang,lihongyang\}@pjlab.org.cn yanjunchi@sjtu.edu.cn}
}

\maketitle

\blankfootnote{$^\ast$ Equal contribution. $^\dagger$ Correspondence author.}

\begin{abstract}
Methods for 3D lane detection have been recently proposed to address the issue of  inaccurate lane layouts in many autonomous driving scenarios (uphill/downhill, bump, \textit{etc.}). Previous work struggled in complex cases due to their simple designs of the spatial transformation between front view and bird's eye view (BEV) and the lack of a realistic dataset. Towards these issues, we present PersFormer: an end-to-end monocular 3D lane detector with a novel Transformer-based spatial feature transformation module. Our model generates BEV features by attending to related front-view local regions with camera parameters as a reference. PersFormer adopts a unified 2D/3D anchor design and an auxiliary task to detect 2D/3D lanes simultaneously, enhancing the feature consistency and sharing the benefits of multi-task learning. Moreover, we release one of the first large-scale real-world 3D lane datasets: OpenLane, with high-quality annotation and scenario diversity. OpenLane contains 200,000 frames, over 880,000 instance-level lanes, 14 lane categories, along with scene tags and the closed-in-path object annotations to encourage the development of lane detection and more industrial-related autonomous driving methods. We show that PersFormer significantly outperforms competitive baselines in the 3D lane detection task on our new OpenLane dataset as well as Apollo 3D Lane Synthetic dataset, and is also on par with state-of-the-art algorithms in the 2D task on OpenLane. The project page is available at \url{https://github.com/OpenPerceptionX/PersFormer_3DLane} 
and OpenLane dataset is provided at \url{https://github.com/OpenPerceptionX/OpenLane}.


\end{abstract}

\section{Introduction}\label{sec: intro}
Autonomous driving is one of the most successful applications for AI algorithms to deploy in recent years.
Modern Advanced Driver Assistance Systems (ADAS) for either L2 or L4 routes provide functionalities such as Automated Lane Centering (ALC) and Lane Departure Warning (LDW), 
where the essential need for perception is a lane detector to generate robust and generalizable lane lines \cite{openpilot}. 
With the prosperity of  
deep learning, lane detection algorithms in the 2D image space has achieved impressive results \cite{tabelini2021keep,liu2021condlanenet,qu2021focus}, where the task is formulated as 
a 2D segmentation problem given front view (perspective) image as input \cite{lee2017vpgnet,pan2018spatial,neven2018towards,abualsaud2021laneaf}.
{However, such a framework to perform lane detection in the perspective view is not applicable 
for industry-level products where complicated scenarios dominate.}

On one side, 
downstream modules as in planning and control \textit{often} require the lane location to be in the form of the orthographic bird's eye view (BEV) instead of a front view representation.
Representation in BEV is for better task alignment with interactive agents (vehicle, road marker, traffic light, \textit{etc}.)  in the environment and multi-modal compatibility with other sensors such as LiDAR and Radar.
The conventional approaches to address such a demand are
either to simply project perspective lanes to ones in the BEV space \cite{wang2014approach,meyer2018deep}, or more elegantly to 
 cast perspective features 
 to BEV by aid of camera in/extrinsic matrices \cite{Garnett_2019_ICCV,guo2020gen,yu2020detecting}. 
The latter solution is inspired by the spatial transformer network (STN) \cite{jaderberg2015spatial} to generate a one-to-one correspondence from the image to BEV feature grids.
%
By doing so, 
the quality of features in BEV
depends solely on the quality of the \textit{corresponding} feature in the front view.
%
The predictions using these outcome features are not adorable as
the blemish of scale variance in the front view, which inherits from the camera's pinhole model, remains.
%

\begin{figure}[t]
    \centering
    \includegraphics[width=\textwidth]{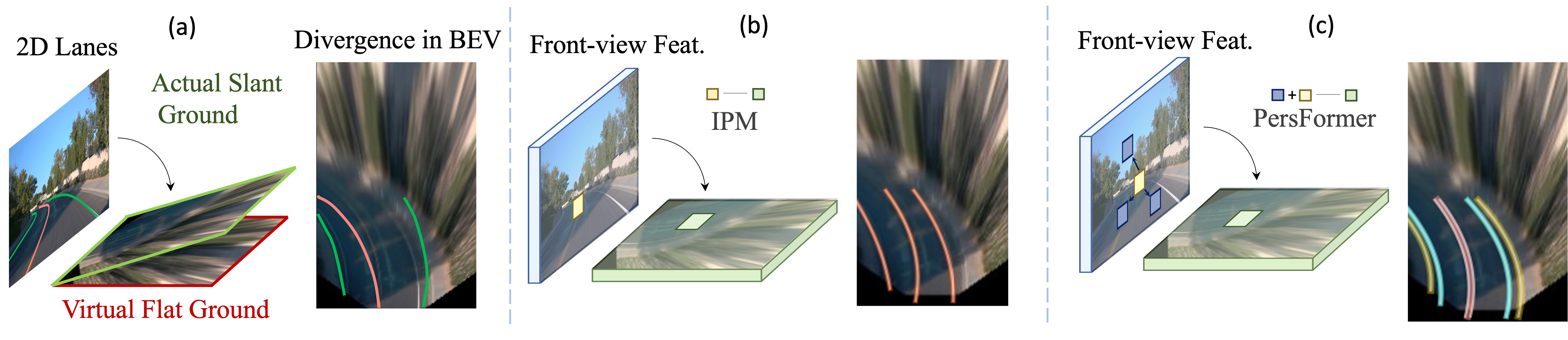}
  \caption{Motivation of performing lane detection from {2D} in (a) to {BEV} in (b); and the superiority of our method in (c) versus (b). Lanes would diverge/converge in projected BEV on planar assumption, and a 3D solution with height to be considered can accurately predict the parallel topology in this case}
  \label{fig:motivation}
\end{figure}

On the other side, the height\footnote{We define the height of lane line $z$ to be the relative height concerning the zero point in the ego vehicle coordinate system $(x,y,z)$ in BEV 3D space. The coordinate of the perspective (front view) 2D space in the image plane is referred to as $(u,v)$.} of lane lines has to be considered when we project perspective lanes into BEV space. As illustrated in Fig. \ref{fig:motivation}, the lanes would diverge/converge in case of uphill/downhill if the height is ignored, leading to improper action decisions as in the planning and control module.
Previous literature \cite{wang2014approach,neven2018towards,su2021structure} inevitably 
hypothesize that lanes in the BEV space lie on a flat ground, \textit{i.e.},  the height of lanes is zero.
The planar assumption does not hold true in most autonomous driving scenarios, \textit{e.g.}, uphill/downhill, bump, crush turn, \textit{etc}. 
{Since the height information is unavailable on public benchmarks or complicated to acquire accurate ground truth,}
3D lane detection is ill-posed.
There are some attempts to address this issue by creating 3D synthetic benchmarks \cite{Garnett_2019_ICCV,guo2020gen}.
%
Their performance still needs improvement in complex, realistic scenarios {nonetheless (c.f. (b-c) in Fig. \ref{fig:motivation})}. 
Moreover, 
the domain adaption between simulation and real data is not well-studied \cite{garnett2020synthetic}.
%

To address these bottlenecks aforementioned, we propose Perspective Transformer, shortened as \textbf{PersFormer}, which has a spatial feature transformation module
to generate better BEV representations for the task.
The proposed framework
unifies 2D/3D lane detection tasks, and substantiates performance on the proposed large-scale realistic 3D lane dataset, \textbf{OpenLane}.

{First, we model the spatial feature transformation as a learning procedure that has an attention mechanism to capture the interaction both among local region in the front view feature and between two views (front view to BEV),}
consequently being able to generate a fine-grained BEV feature representation.
Inspired by \cite{vaswani2017attention,carion2020end}, we construct a Transformer-based module to realize this, while the deformable attention mechanism \cite{zhu2021deformable} is adopted to remarkably reduce the computational memory requirement and 
%
{dynamically adjust keys through the cross-attention module to capture prominent feature among the local region.}
%
{Compared with direct 1-1 transformation via Inverse Perspective Mapping (IPM), the resultant features would be more representative and robust as it attends to the surrounding local context and aggregates relevant information.}
%
{We further aim at unifying 2D and 3D lane detection tasks to benefit from the co-learning optimization.}
%
%
%
Second, we release the first real-world, large-scale 3D lane dataset and corresponding benchmark, OpenLane, to support research into the problem.
OpenLane contains 200,000 annotated frames and over 880,000 lanes - each with one of 14 category labels (single white dash, double yellow solid, left/right curbside, \textit{etc}.), which exceeds all of the existing lane datasets.
It also has some distinguishing elements such as scenes, weather, and closed-in-path-object (CIPO) for other research topics in autonomous driving.

\textbf{The main contributions of our work are three-fold:}
\textbf{1)}
Perspective Transformer, a novel Transformer-based architecture to realize spatial transformation of features;
\textbf{2)} An architecture to simultaneously unify 2D and 3D lane detection, which is feasibly needed in the application. %
Experiments show that our PersFormer outperforms state-of-the-art 3D lane detection algorithms;
\textbf{3)} The OpenLane dataset, the first large-scale realistic 3D lane dataset with high-quality labeling and vast diversity. The dataset, baselines, as well as the whole suite of codebase, is released to facilitate the research in this area.

\section{Related Work}\label{sec: related work}

\textbf{Vision Transformers in Bird's-Eye-View (BEV).}
Projecting features to BEV and performing downstream tasks in it has become more dominant and ensured better performance recently \cite{patric2021blog}. 
Compared with conventional CNN structure, the cross attention scheme in Vision Transformers \cite{vaswani2017attention,dosovitskiy2021an,carion2020end,liu2021swin,zhu2021deformable} 
is 
{naturally introduced to serve as a learnable transformation of features across different views in an elegant spirit \cite{patric2021blog}. 
}
%
%
{Instead of simply projecting features via IPM, the successful application of Transformers in view transformation has demonstrated great success in various domains, including 3D object detection \cite{yin2020lidar,wang2022detr3d,guan2022m3detr,li2022bevformer}, prediction \cite{gao2020vectornet,gu2021densetnt,ngiam2021scene}, planning \cite{prakash2021multi,chitta2021neat}, \textit{etc}. 
}

Previous work \cite{Garnett_2019_ICCV,yang2021projecting,wang2022detr3d,saha2021translating,can2021structured} bring the BEV philosophy into pipeline, and yet they do not consider attention mechanism and/or 3D vision geometry (in this case, camera parameters).
For instance, 3D-LaneNet \cite{Garnett_2019_ICCV} is set up with camera in/extrinsic matrices; the IPM process generates a virtual BEV representation from front view features.
{DETR3D} \cite{wang2022detr3d} also considers camera geometry 
and formulates a learnable 3D-to-2D query search with attention scheme.
However, there is no explicit BEV modelling for robust feature representation; 
{the aggregated features might not be properly represented in 3D space.}
{To address these shortcomings, our proposed PersFormer takes into account both the effect of camera parameters 
to generate BEV features 
and the convenience of cross-attention mechanism to model view transformation, 
achieving better feature representation in the end.}

\noindent\textbf{Lane Detection Benchmarks.}
A large-scale, diverse dataset with high-quality annotation is a pivot for lane detection. 
Along with the progress of lane detection approaches, numerous datasets have been proposed \cite{lee2017vpgnet,huang2019apolloscape,yu2020bdd100k,tusimple2017,llamas2019,pan2018spatial,xu2020curvelane,ChenICME22}.
However, they usually fit into one or the other lane detection scenario.
%
%
%
Tab. \ref{tab: datasets_comparison} depicts more details of the existing benchmarks and their comparison with our proposed OpenLane dataset. 
OpenLane is the first large-scale, realistic 3D lane dataset. It equips with a  wide span of diversity in both data distribution and task applicability.

\noindent\textbf{3D Lane Detection.} As discussed in Section \ref{sec: intro}, planar assumption does not always reserve in some cases, \textit{i.e.}, 
uphill/downhill, bump. 
Several approaches \cite{nedevschi20043d,benmansour2008stereovision,bai2018deep} utilize multi-modal or multi-view sensors, such as a stereo camera or LiDAR, to get the 3D ground topology. 
%
However, these sensors have shortages of high cost in hardware and computation resources, confining their practical applications. 
Recently, some monocular methods \cite{Garnett_2019_ICCV,guo2020gen,jin2021robust,liu2022learning} take a single image and employ IPM to predict lanes in 3D space.
3D-LaneNet \cite{Garnett_2019_ICCV} is the pioneering work in this domain with one simple end-to-end neural network, which adopts STN \cite{jaderberg2015spatial} to accomplish the spatial projection of features. 
Gen-LaneNet \cite{guo2020gen} builds on top of 3D-LaneNet and designs a two-stage network for decoupling the segmentation encoder and 3D lane prediction head. 
These two approaches~\cite{Garnett_2019_ICCV,guo2020gen} suffer from improper feature transformation and unsatisfying performance in curving or crush turn cases.
Confronted with the issues above, 
we bring in PersFormer to provide better feature representation 
and
optimize 
anchor design to unify 2D and 3D lane detection simultaneously.

\section{Methodology}\label{sec: algorithm}
\begin{figure}[t]
\centering
  \includegraphics[width=0.9\textwidth]{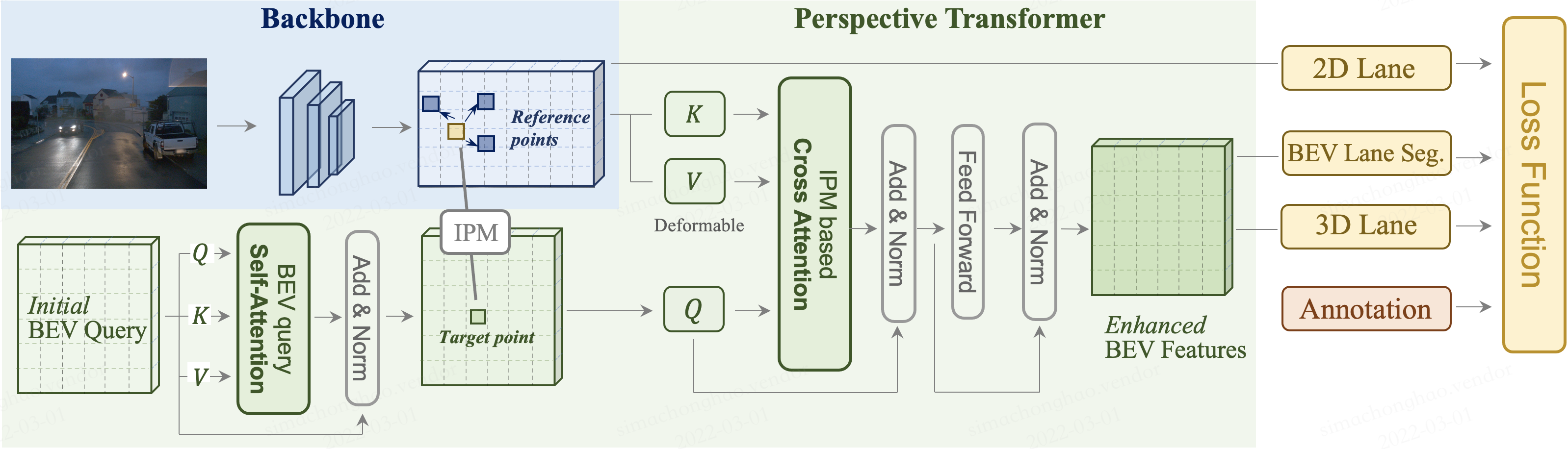}
  \caption{Our proposed PersFormer pipeline. The core is to learn a spatial feature transformation from front view to BEV space so that the generated BEV features at target point would be more representative by attending local context around reference point. 
  %
  PersFormer consists of the self-attention module to interact with its own BEV queries; the cross-attention module that takes the key-value pair from the IPM-based front view features to generate fine-grained BEV feature
  }
  \label{fig:pipeline}
\end{figure}

In this section, we propose PersFormer, a unified 2D/3D lane detection framework with Transformer. 
We first describe the problem formulation, followed by an introduction to the overall structure in Section \ref{sec: alg - pipeline}. 
In Section \ref{sec: alg - transformer}, we present Perspective Transformer, an explicit feature transformation module
{from front view to BEV space by the aid of camera parameters}.
In Section \ref{sec: alg - anchor}, we give details on the anchor design to unify 2D/3D tasks
and in Section \ref{sec: alg - loss} we further elaborate on the auxiliary task and loss function to finalize our training strategy
.

%
\textbf{Problem Formulation.}
Given an input image $I_{org} \in \mathbb{R}^{H_{org} \times W_{org}}$, the goal of PersFormer is to predict a collection of 3D lanes $L_{3D} = \{ l_1, l_2, \dots, l_{N_{3D}} \}$ and 2D lanes ${L_{2D} = \{ l_1, l_2, \dots, l_{N_{2D}} \}}$, 
where $N_{3D}, N_{2D}$ are the total number of 3D lanes in the pre-defined BEV range and 2D lanes in the original image space (front view) respectively.
Mathematically, each 3D lane $l_d$ is represented by an ordered set of 3D coordinates:
\begin{equation}
    l_d = \big[(x_{1}, y_{1}, z_{1}), (x_{2}, y_{2}, z_{2}), \dots, (x_{N_d}, y_{N_d}, z_{N_d}) \big],
    \label{eqn: 3d_lane_represent}
\end{equation}
where $d$ is the lane index, and $N_d$ is the max number of sample points of this lane. The form of 2D lane is represented similarly with 2D coordinate $(u,v)$ accordingly.
Each 
lane has a categorical attribute ${c_{\text{3D/2D}}}$, 
indicating the type of this lane (\textit{e.g.}, 
single-white dash line). 
%
Also, for each point in a single 2D/3D lane, there exists an attribute property indicating whether the point is visible or not, denoted by $\textbf{vis}_{\text{fv/bev}}$ as a vector for the lane.
%


\subsection{Approach Overview}\label{sec: alg - pipeline}

%
%
%
The overall structure, 
as illustrated in Fig. \ref{fig:pipeline}, consists of three parts: the backbone, the Perspective Transformer, and lane detection heads.
%
The backbone takes the resized image as input 
and generates multi-scale front view features, 
where the popular ResNet variant \cite{tan2019efficientnet} is adopted.
Note that these features might suffer from the defect of scale variance, occlusion,\textit{ etc.} - residing from the inherent feature extraction in the front view space.
The Perspective Transformer takes the front view features as input and generates BEV features by the aid of camera intrinsic and extrinsic parameters. 
%
Instead of simply projecting the one-to-one feature correspondence from the front view to BEV, we introduce Transformer to attend local context 
and aggregate surrounding features to form a robust representation in BEV.
By doing so, we learn the inverse perspective mapping from front view to BEV in an elegant manner with Transformer.
Finally, the lane detection heads are responsible for predicting 2D/3D coordinates as well as lane types.
The 2D/3D detection heads are referred to as LaneATT \cite{tabelini2021keep} and 3D-LaneNet \cite{Garnett_2019_ICCV}, with modification on the structure and anchor design.
%
%
%

\subsection{Proposed Perspective Transformer}\label{sec: alg - transformer}

%

%

%
We present Perspective Transformer, a spatial transformation method that combines camera parameters and data-driven learning procedures.
The general idea of Perspective Transformer is
{to use the coordinates transformation matrix from IPM as a reference to generate BEV feature representation},
by attending related region (local context) in front view feature.
On the assumption that the ground is flat and the camera 
parameters are given, a classical IPM approach calculates a set of coordinate mapping from front-view to BEV, where the BEV space is defined on the flat ground (see \cite{Hartley2004}, Section 8.1.1).
%
Given a point $p_{\text{fv}}$ with its coordinate $(u,v)$ in the front-view feature $F_{\text{fv}} \in \mathbb{R}^{H_{\text{fv}} \times W_{\text{fv}} \times C}$, IPM maps the point $p_{\text{fv}}$ to the corresponding point $p_{\text{bev}}$ in BEV, where $(x,y)$ is the coordinate in the BEV space $\mathbb{R}^{H_{\text{bev}} \times W_{\text{bev}} \times C}$. 
The transform is achieved with camera in/extrinsic and can be represented mathematically as:
%
%
\begin{equation}
    \begin{pmatrix}
    x \\
    y \\
    0
    \end{pmatrix} = \alpha_{f2b} \cdot R_{\theta} \cdot K^{-1} \cdot
    \begin{pmatrix}
    u \\
    v \\
    1
    \end{pmatrix} + 
    \begin{pmatrix}
    0 \\
    0 \\
    -h
    \end{pmatrix}, \label{Enq: transform}
\end{equation}
where $\alpha_{f2b}$ implies the scale factor between front-view and BEV, $R_{\theta}$ denotes the pitch rotation matrix from extrinsic, $K$ is the intrinsic matrix, and $h$ stands for camera height.
%
Such a transformation in Eqn.(\ref{Enq: transform}) enframes a strong prior 
on the attention unit in PerFormer to generate more representative BEV features.

The architecture of Perspective Transformer is inspired by popular approaches such as DETR \cite{carion2020end},  and consists of the self-attention module and cross-attention module (see Fig. \ref{fig:pipeline}).
We differentiate from them in that the queries are not implicitly updated. However, instead, they are piloted by an explicit meaning - the physical location to detect objects or lanes in BEV.
%
{In the \textbf{self-attention} module, the output $ Q_{\text{bev}} $ descends from the triplet (key, value, query) input through their interaction.}
The formulation of such a self-attention can be described as:
\begin{equation}
    Q_{\text{bev}} = \texttt{softmax} \bigg( \frac{QK^\top}{\sqrt{d_k}} \bigg) V,
\end{equation}
where $K,Q,V \in \mathbb{R}^{(H_{\text{bev}} \times W_{\text{bev}} \times C)}$ are the same query that is pre-defined in BEV, $\sqrt{d_k}$ is the dimensional normalized factor.

In the \textbf{cross-attention} module, the input query $Q^{'}_{\text{bev}}$ is the outcome of several additional layers feeding the self-attention output $Q_{bev}$ as input.
{Note that $Q^{'}_{\text{bev}}$ is an explicit feature representation as to which part in BEV should be paid more attention since the generation of queries is location-sensitive in BEV.}
This is quite different compared with 
queries 
{that do not consider view transformation} 
in most Vision Transformers \cite{wang2022detr3d,guan2022m3detr,zhu2021deformable}. 
Furthermore,
the intuition behind employing Transformer to map features from front view to BEV is that such an attention mechanism would automatically attend which part of features contribute \textit{most} towards the {target point (query) in the destination view.}
%
The direct feature transformation would suffer from camera parameter noise or scale variance issues, as discussed and illustrated in Section \ref{sec: intro}.
Note that the naive Transformer cannot be applied directly since the number of key-value pairs is huge and thus be confined by computational burden.
Inspired by Deformable DETR \cite{zhu2021deformable}, we attend partial key-value pairs around the local region in a learnable manner to save cost and improve efficiency.

\begin{figure}[t]
\begin{minipage}[b]{.49\textwidth}
  \centering
  \includegraphics[width=\linewidth]{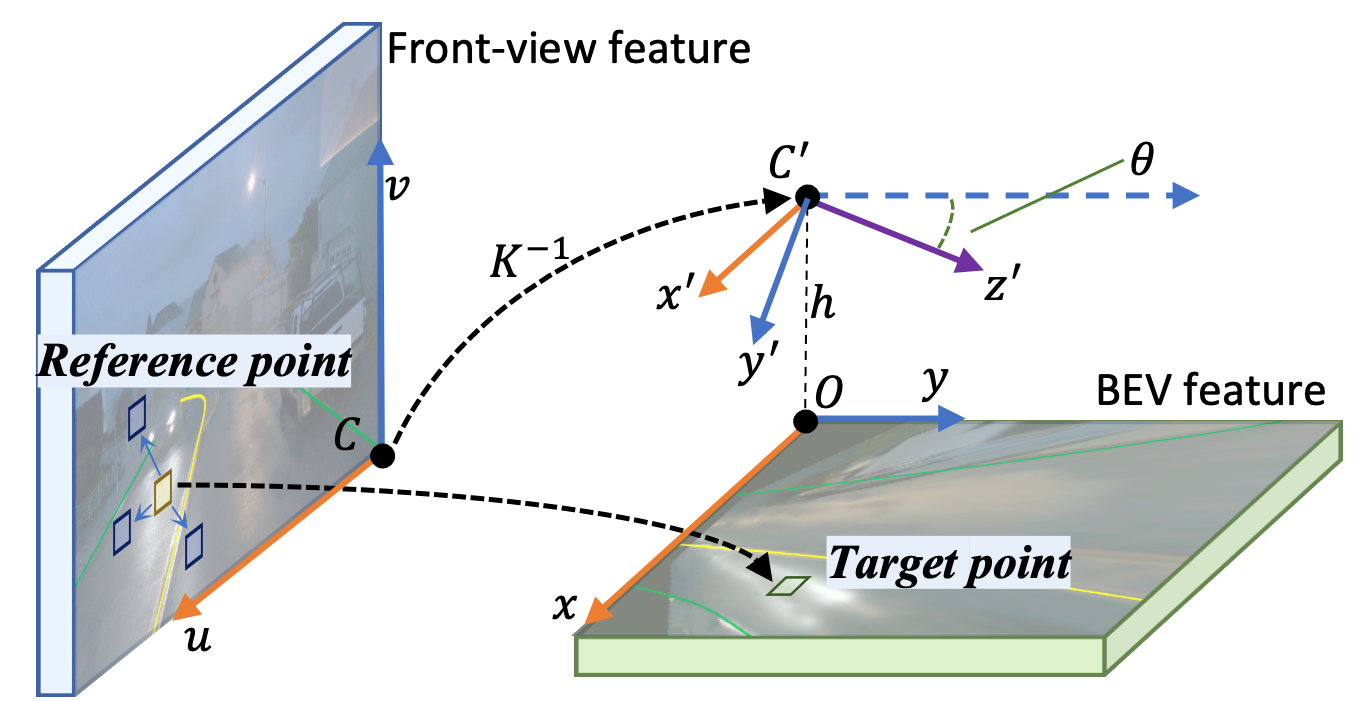}
  \caption{
 Generation of keys in the cross attention. 
 Point $(x,y)$ in BEV space casts the corresponding point $(u,v)$ in front view through intermediate state $(x^{'},y^{'})$; by learning offsets,
 the network learns target-reference points mapping
 {from green rectangles to yellow and related blue rectangles}
 as keys to Transformer
 }
  \label{fig:transformer} 
\end{minipage}
~~~~
\begin{minipage}[b]{.49\textwidth}
  \centering
  \includegraphics[width=\linewidth]{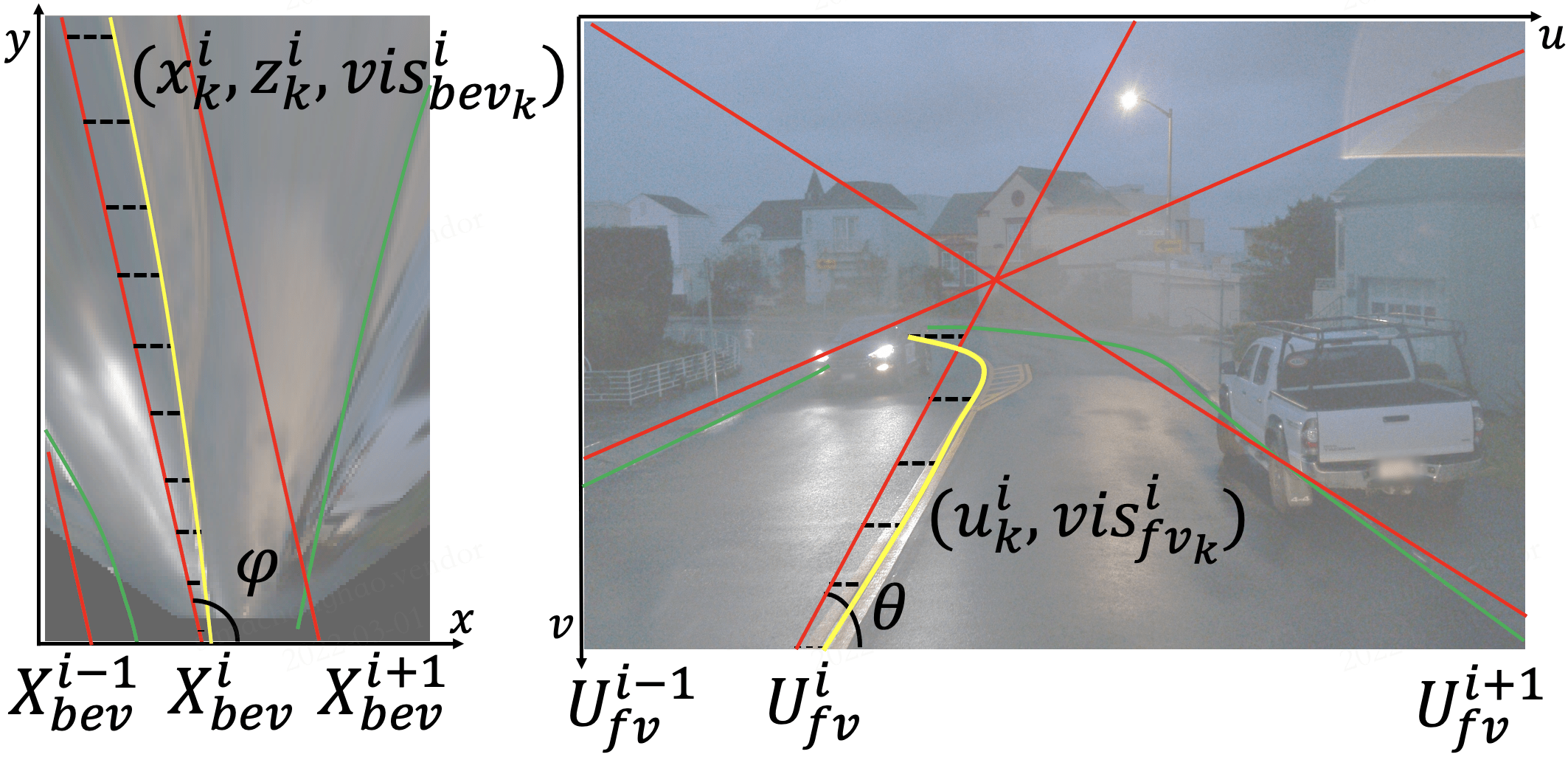}
   \caption{Unifying anchor design in 2D and 3D. 
  We first put curated anchors (red) in the BEV space (left), then project them to the front view (right). Offset $x_k^{i}$ and $u_k^{i}$ (dashed line) are predicted to match ground truth (yellow and green) to anchors. The correspondence is thus built, and features are optimized together
  }
  \label{fig:anchor} 
\end{minipage}
\end{figure}

Fig. \ref{fig:transformer} depicts the feature transformation process and the generation of key-value pairs in cross-attention.
Specifically, given a query point $(x,y)$ in the target BEV map $Q^{'}_{\text{bev}}$, we project it to the corresponding point $(u,v)$ in the  front view via Eqn.(\ref{Enq: transform}). As does similarly in \cite{zhu2021deformable}, we learn some offsets based on point $(u,v)$ 
to generate a set of most related points around it. 
{These learned points, together with $(u,v)$ are defined as \textit{reference points}.}
They contribute most to the query point $(x,y)$,  defined as \textit{target point}, in BEV-space.
{The reference points} serve as {the surrounding context in the local region} that contributes most to the feature representation from perspective view to BEV space.
%
They are the desired keys we try to find, and their features are values for the cross attention module.
%
%
Note that the initial locations of reference points from IPM are used as preliminary locations for the coordinate mapping; 
the location 
are adjusted gradually during the learning procedure, which is the core role of Deformable Attention.

%
As a result, the output of the cross-attention module can be formulated as:
\begin{equation}
    F_{\text{bev}} = \texttt{DeformAttn}(Q^{'}_{\text{bev}}, F_{\text{fv}}, p_{\text{fv2bev}}),
\end{equation}
where $F_{\text{bev}} \in \mathbb{R}^{(H_{\text{bev}} \times W_{\text{bev}} \times C)}$ is the final desired features for the subsequent 3D head to get lane predictions, 
$Q^{'}_{\text{bev}}$ denotes the input queries, 
$F_{\text{fv}} \in \mathbb{R}^{(H_{\text{fv}} \times W_{\text{fv}} \times C)}$ indicates the front view features from backbone, 
and $p_{\text{fv2bev}}$ is 
the IPM-inited coordinate mapping from front view to BEV space. 
Considering $F_{\text{fv}}$ and $p_{\text{fv2bev}}$ with the deformable unit, we get the explicit transformed BEV feature $F_{\text{bev}}$.
%


To sum up, Perspective Transformer extracts front-view features among the reference points to construct representative BEV features.
%
As demonstrated in Section \ref{sec: experiments}, such a feature transformation in an aggregation spirit via Transformer
is proven to perform better than a direct IPM-based projection across views. 
%
%

\subsection{Simultaneous 2D and 3D Lane Detection}
\label{sec: alg - anchor}
%
%
%

Although the main focus in this paper lies in 3D detection, we formulate the  PersFormer framework to detect 2D and 3D lanes in one shot. On one side, 2D lane detection in the perspective view still draws interest in the community as part of the general high-level vision problems \cite{abualsaud2021laneaf,tabelini2021keep,liu2021condlanenet,qu2021focus}; on the other side, unifying 2D and 3D tasks are naturally feasible since the BEV features to predict 3D outputs descend from the counterpart in the 2D branch. An end-to-end unified framework would leverage features 
and benefit from the co-learning optimization process as proven in most multi-task literature \cite{liang2019multi,vandenhende2021multi,kumar2021omnidet}.  

\textbf{Unified anchor design.} 
Since our method is anchor-based detection, the core issue to achieve the unified framework is to integrate anchors in both 2D and 3D.
Unfortunately, anchors in these two domains usually do not share similar distribution. 
For example, the popular 2D approach LaneATT \cite{tabelini2021keep} settles too many anchors, spanning different directions in the image; while the 
recent 3D work Gen-LaneNet \cite{guo2020gen} puts too few anchors, which are parallel and sparse in BEV.
%
Based on these observations, 
we thereby
design anchors
such that 
the redesigned anchors could leverage the network to optimize shared features across \textit{two} domains.
%
%
We start with several groups of anchors (here, the group number is set to 7)  sampled with different incline angles in the BEV space and then projected to the front view.
Fig. \ref{fig:anchor} elaborates on the integration of 2D and 3D anchors.
Below we describe how the lane line is modeled via anchors.

\textbf{3D anchor design.} 
To match ground truth lanes tightly, 
the anchors are placed approximately longitudinal along \textit{x}-axis, with an incline angle $\varphi$.
As denoted in Fig. \ref{fig:anchor}(left), 
{the initial line (equally spaced) with staring  position along \textit{x}-axis is denoted by $X^i_{\text{bev}}$} for each anchor $i$.
Similar to anchor regression in object detection, the network predicts the relative offset $\textbf{x}^i$ w.r.t. the initial position
$X^i_{\text{bev}}$; hence the resultant lane prediction along \textit{x}-axis 
is $(\textbf{x}^i + X^i_{\text{bev}})$. 
As indicated in Eqn.(\ref{eqn: 3d_lane_represent}), each lane is represented as a number of $N_d$ points. The prediction head  generates three vectors related to lane shape as follows: 

\begin{equation}
    (\mathbf{x}^i, \mathbf{z}^i, \mathbf{vis}^i_{\text{bev}}) = \{ (x^{(i,k)}, z^{(i,k)}, \text{vis}^{(i,k)}_{\text{bev}} ) \}^{N_d}_{k=1}
\end{equation}
where $\textbf{z}^i$ is the lane height in 3D sense,  the binary $\text{vis}^{(i,k)}_{\text{bev}}$ denotes the visibility of each location $k$  in  lane $i$, which controls the endpoint or length of a lane. 
%
%
Note that the lane position along \textit{y}-axis does not need to be predicted since each \textit{y} value of the $N_d$ samples in a lane is pre-defined - we predict the $x^{(i,k)}$ value at the corresponding (fixed) $y$ location. 
To sum up, the description of a lane's location in the world coordinate system is denoted as
 $(\textbf{x}^i + X^i_{\text{bev}}, \textbf{y}, \textbf{z}^i)$.


%
\textbf{2D anchor design.} The anchor description and prediction are similar to those defined in 3D view, 
except that the $(u, v)$ is in 2D space and there is no height (see Fig. \ref{fig:anchor}({right})).
We omit the detailed notations for brevity.
%
%
It is worth mentioning that each 3D anchor $X^i_{\text{bev}}$ with an incline angle $\varphi$ corresponds to a specific 2D anchor $U^i_{\text{fv}}$ with the incline angle $\theta$; the connection is built via the projection in Eqn.(\ref{Enq: transform}).
We achieve the goal of unifying 2D and 3D tasks simultaneously by setting the \textit{same} set of anchors. Such a design would optimize features together and features being more aligned and representative across views.
%
%


\subsection{Prediction Loss}\label{sec: alg - loss}

\textbf{Binary Segmentation under BEV.} 
As do in many preceding work \cite{wei2016convolutional,newell2016stacked,huang2019apolloscape}, adding more intermediate supervision into the network training would
{boost the performance of network}.
Since lane detection belongs to image segmentation and requires general large resolution, 
we concatenate a U-Net structure \cite{ronneberger2015u} head on top of the generated BEV features. 
%
Such an auxiliary task is to predict lanes in BEV, but instead in a conventional 2D segmentation manner, aiming for better feature representation for the main task.
The ground truth $S_{gt}$ 
is a binary segmentation map projected from 3D lane ground truth to the BEV space.
%
The prediction output is denoted by $S_{ \text{pred}}$ and owns the same size as $S_{ \text{gt} }$.

%
\textbf{Loss function.} 
Equipped with the anchor representation and segmentation head aforementioned, we summarize the overall loss.
Given an image input and its ground truth labels, 
it finally computes a sum of all anchors' loss;
%
the loss is a combination of the 2D lane detection, 3D lane detection and intermediate segmentation with learnable weights $(\alpha, \beta, \gamma)$ accordingly:
%
\begin{equation}\label{eqn:total_loss}
        \mathcal{L} = 
        \sum_i  
        \alpha \mathcal{L}_{ \text{2D}  }( c^i_{  \text{2D}   }, \mathbf{u}^i, \mathbf{vis}^i_{\text{fv}} ) + \beta 
        \mathcal{L}_{ \text{  3D } } ( c^i_{ \text{3D} }, \mathbf{x}^i, \mathbf{z}^i, \mathbf{vis}^i_{\text{bev}} ) + \gamma \mathcal{L}_{ \text{seg}} (S_{\text{pred}}),
\end{equation}
where $c^i_{( \cdot )} $ is the predicted lane category in 2D and 3D domain respectively. The loss input above shows the prediction part only; we omit the ground truth notation for brevity.
The loss of lane category classification for the 2D/3D task is the cross-entropy; the loss of lane shape regression is the $l_1$ norm; the loss of lane visibility prediction is the binary cross-entropy loss.
The loss of the auxiliary task is a binary cross-entropy loss between two segmentation maps.

\section{OpenLane: A Large-scale Realistic 3D Lane Benchmark}\label{sec: dataset}
\begin{table}[t]
\caption{Comparison of OpenLane with existing benchmarks.
%
``Avg. Length" denotes the average time duration of segments. ``Inst. Anno." indicates whether lanes are annotated instance-wise (c.f. semantic-wise). ``Track. Anno." implies if a lane 
has a unique tracking ID. Numbers in `\#Frames' are the number of annotated frames / total frames respectively.
Details of ``Scenario" can be found in Appendix
%
}
\centering
\label{tab: datasets_comparison}
\scalebox{0.82}{
\begin{tabular}{p{0.24\textwidth}>{\centering}p{0.13\textwidth}>{\centering}p{0.15\textwidth}>{\centering}p{0.1\textwidth}>{\centering}p{0.1\textwidth}>{\centering}p{0.1\textwidth}>{\centering}p{0.1\textwidth}>{\centering}p{0.1\textwidth}>{\centering\arraybackslash}p{0.12\textwidth}}
\toprule
Dataset         & \#Segments & \#Frames           & \begin{tabular}[c]{@{}c@{}}Avg.\\ Length\end{tabular} & \begin{tabular}[c]{@{}c@{}}Inst.\\ Anno.\end{tabular} & \begin{tabular}[c]{@{}c@{}}Track.\\ Anno.\end{tabular} & \begin{tabular}[c]{@{}c@{}}Max\\ \#Lanes\end{tabular} & \begin{tabular}[c]{@{}c@{}}Line\\ Category\end{tabular} & Scenario      \\ \midrule
Caltech Lanes \cite{aly2008real}   & 4          & 1224/1224          & -                                                     & \ding{51}                                                   & \ding{55}                                                     & 4                                                     & -                                                       & Easy          \\
TuSimple \cite{tusimple2017}        & 6.4K       & 6.4K/128K          & 1s                                                    & \ding{51}                                                   & \ding{55}                                                     & 5                                                     & -                                                       & Easy          \\
3D Synthetic \cite{guo2020gen}    & -          & 10K/10K            & -                                                     & \ding{51}                                                   & -                                                      & 6                                                     & -                                                       & Easy          \\
VIL-100 \cite{zhang2021vil}         & 100        & 10K/10K            & 10s                                                   & \ding{51}                                                   & \ding{55}                                                     & 6                                                     & 10                                                      & Medium        \\
VPG \cite{lee2017vpgnet}             & -          & 20K/20K            & -                                                     & \ding{55}                                                    & -                                                      & -                                                     & 7                                                       & Medium        \\
OpenDenseLane \cite{ChenICME22}      &1.7K        & 57K/57K         & -                                                     & \ding{51}                                                   & \ding{55}                                                     & -                                                    & 4                                                       & Medium        \\
LLAMAS \cite{llamas2019}           & 14         & 79K/100K           & -                                                     & \ding{51}                                                   & \ding{55}                                                     & 4                                                     & -                                                       & Easy          \\
ApolloScape \cite{huang2019apolloscape}    & 235        & 115K/115K          & 16s                                                   & \ding{55}                                                    & \ding{55}                                                     & -                                                     & 13                                                      & Medium        \\
BDD100K \cite{yu2020bdd100k}        & 100K       & 100K/120M          & 40s                                                   & \ding{55}                                                    & \ding{55}                                                     & -                                                     & 11                                                      & Medium        \\
CULane \cite{pan2018spatial}          & -          & 133K/133K          & -                                                     & \ding{51}                                                   & -                                                      & 4                                                     & -                                                       & Medium        \\
CurveLanes \cite{xu2020curvelane}      & -          & 150K/150K          & -                                                     & \ding{51}                                                   & -                                                      & 9                                                     & -                                                       & Medium        \\
ONCE-3DLanes \cite{yan2022once}      & -          & 211K/211K          & -                                                     & \ding{51}                                                   & -                                                      & 8                                                     & -                                                       & Medium        \\
\rowcolor{lightgray}
\textbf{OpenLane} & 1K         & \textbf{200K/200K} & 20s                                                   & \ding{51}                                                   & \ding{51}                                                    & \textbf{24}                                           & \textbf{14}                                             & \textbf{Hard} \\ \bottomrule
\end{tabular}}
\end{table}
%
%
\subsection{Highlights over Previous Benchmarks} \label{sec: dataset - comparison}
%
OpenLane is the \textit{first} real world 3D lane dataset and the \textit{largest} scale to date compared with existing benchmarks.
We construct OpenLane on top of the influential Waymo Open dataset \cite{sun2020scalability}, following the same data format and evaluation pipeline - leveraging existent practice in the community so that users would not handle additional rules for a new benchmark.
Tab. \ref{tab: datasets_comparison} compares OpenLane with existing counterparts in various aspects. In short, OpenLane owns 200K frames and over 880K carefully annotated lanes, 33\% and 35\% more compared with existing largest lane dataset CurveLanes \cite{xu2020curvelane} respectively, with rich annotations.

\begin{figure}[t!]
    \centering
    \begin{minipage}[c]{0.5\textwidth}
      \includegraphics[width=\textwidth]{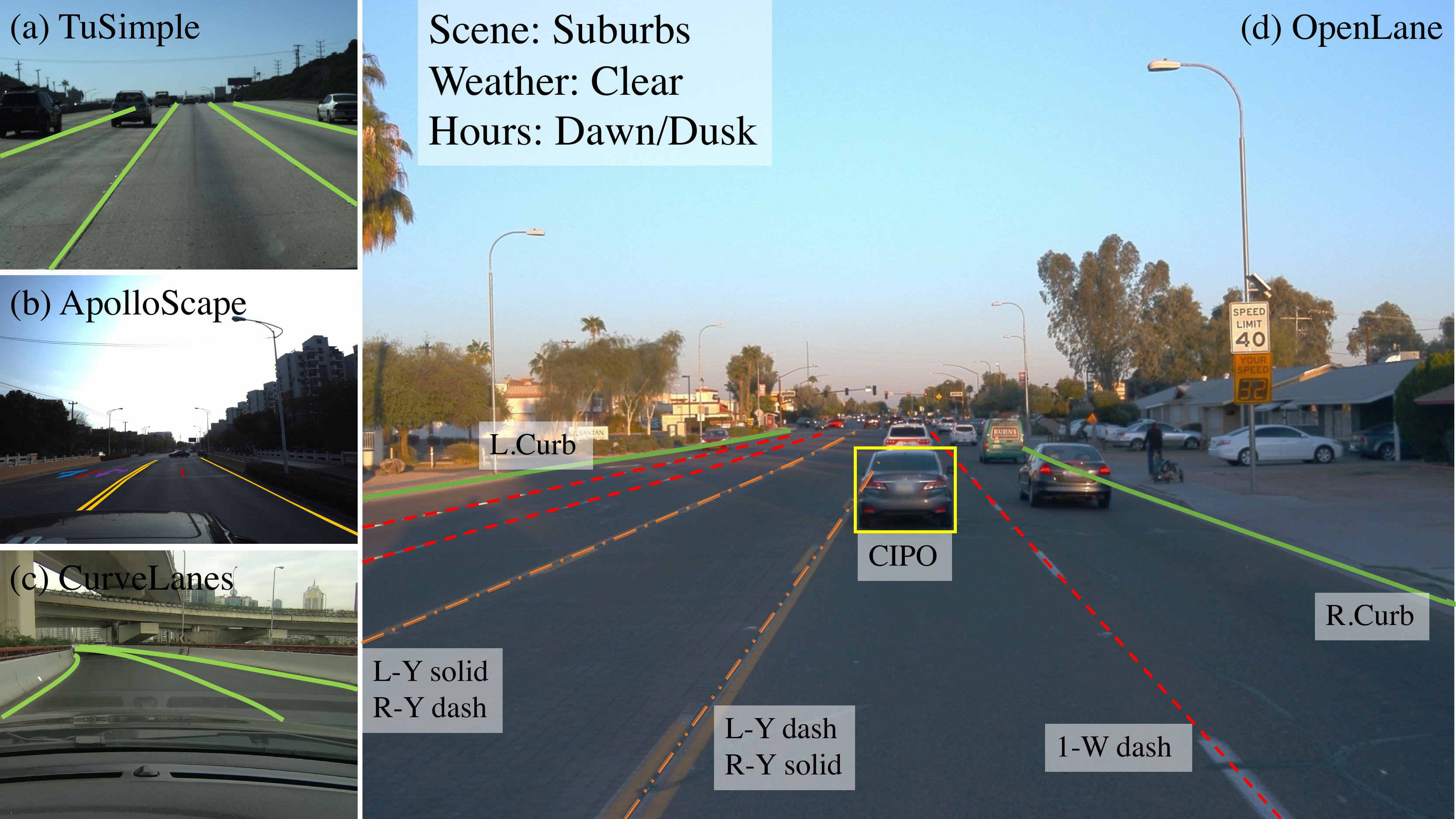}
    \end{minipage} ~ 
    \begin{minipage}[c]{0.34\textwidth}
      \caption{Annotation samples of OpenLane compared with other lane datasets. OpenLane is challenging with more lane categories per frame in average and has rich labels including scene, weather, hours, CIPO}
    \label{fig: dataset_comparison}
    \end{minipage}
\end{figure}

We annotate all the lanes in each frame, including those in the \textit{opposite} direction if no curbside exists in the middle. Due to the  complicated lane topology, \textit{e.g.}, intersection/roundabout, one frame could contain 
as many as
\textbf{24} lanes in OpenLane. Statistically, about 25\% frames of OpenLane have more than 6 lanes, which exceeds the maximum number in most lane datasets. 
\textbf{14} lane categories are annotated alongside to cover a wide range of lane types in most scenarios, including road edges. 
Double yellow solid lanes, single white solid and dash lanes take up almost 90\%
of total lanes. 
This is imbalanced, and yet it falls into a long-tail distribution
problem, which is common in realistic scenarios.
%
%
In addition to the lane detection task, we also annotate: (a) scene tags, such as weather and locations; (b) the closest-in-path object (CIPO), which is defined as the most concerned target w.r.t. ego vehicle; such a tag is quite pragmatic for subsequent modules as in planning/control, besides a whole set of objects from perception. 
An annotation example is provided in Fig. \ref{fig: dataset_comparison}(d), along with some typical samples in existing 2D lane datasets in Fig. \ref{fig: dataset_comparison}(a-c).
The detailed statistics, annotation criterion and visualization  
can be found in Appendix.
%
%
%
\subsection{Generation of High-quality Annotation} \label{sec: dataset - gen}
Building a real-world 3D lane dataset has challenges mainly in an accurate localization system and occlusions.
{We compare several popular sensor datasets \cite{chang2019argoverse,sun2020scalability,caesar2020nuscenes} by projecting 3D object annotations to image planes and constructing 3D scene maps using both learning-based \cite{teed2021droid} or SLAM algorithms \cite{liosam2020shan,lvisam2021shan}.}
The reconstruction precision and scalability of Waymo Open Dataset \cite{sun2020scalability} outperforms other candidates, leading to employing it as our basis.

Primarily, we generate the necessary high-quality 2D lane labels. 
They contain the final annotations of tracking ID, category, and 2D points ground truth.
Then for each frame, the point clouds are first filtered with the original 3D object bounding boxes and then projected back into the corresponding image. 
We further keep those points related to 2D lanes only with a certain threshold.
However, the output directly after a static threshold filtering could lead to an unsatisfying ground truth due to the perspective scaling issue.
To solve this and keep the slender shape of lanes, we use the filtered point clouds to interpolate the 3D position for each point in 2D annotations. 
Afterward, with the help of the localization system, 3D lane points in frames within a segment could be spliced into long, high-density lanes.
This process could bring some unreasonable parts into the current frame; thus, points in one lane whose 2D projections are higher than the ending position of its 2D annotation are labeled as invisible.
A smoothing step is ultimately deployed to filtrate any outliers and generate the 3D labeling results.
We omit some technical details, such as how to deal with a large U-turn during smoothing, and we refer the audience to Appendix.

\section{Experiments}\label{sec: experiments}

%
We examine PersFormer on two 3D lane benchmarks, the newly proposed {real-world} OpenLane dataset, and the {synthetic} Apollo dataset. 
%
%
%
%
%
For both 3D lane datasets, we follow the evaluation metrics designed by Gen-LaneNet \cite{guo2020gen}, with additional category accuracy on OpenLane dataset.
For the 2D task, the classical metric in CULane \cite{pan2018spatial} is adopted.
We put correlated details in Appendix.

\begin{table*}[t]
\centering

\begin{minipage}[t]{\textwidth}
\caption{Comparison with other open-sourced 3D methods on OpenLane. PersFormer achieves the best F-Score on the entire validation set and every scenario set}
\centering
\label{tab: OpenLane 3d cases results}
\scalebox{0.8}{
\begin{tabular}{p{0.25\textwidth}>{\centering}p{0.12\textwidth}>{\centering}p{0.12\textwidth}>{\centering}p{0.12\textwidth}>{\centering}p{0.12\textwidth}>{\centering}p{0.12\textwidth}>{\centering}p{0.13\textwidth}>{\centering\arraybackslash}p{0.13\textwidth}}
\toprule
Method      & All  & \begin{tabular}[c]{@{}c@{}}Up \&\\ Down\end{tabular} & Curve & \begin{tabular}[c]{@{}c@{}}Extreme\\ Weather\end{tabular} & Night & Intersection & \begin{tabular}[c]{@{}c@{}}Merge \&\\ Split\end{tabular} \\ \midrule
3D-LaneNet \cite{Garnett_2019_ICCV}  & 44.1 & 40.8  & 46.5  & 47.5   & 41.5  & 32.1         & 41.7   \\
Gen-LaneNet \cite{guo2020gen} & 32.3 & 25.4  & 33.5  & 28.1  & 18.7  & 21.4         & 31.0                                                     \\

\rowcolor{lightgray}
PersFormer (ours)        & \textbf{50.5}      &                 \textbf{42.4}                                     &   \textbf{55.6}    &                         \textbf{48.6}                                  &  \textbf{46.6}     &       \textbf{40.0}       &                    \textbf{50.7}                                      \\ \bottomrule
\end{tabular}
}

\end{minipage}

\bigskip

\begin{minipage}[t]{\textwidth}

\caption{Comparison with state-of-the-art 2D method on OpenLane. The result from the 2D head of PersFormer also achieves competitive performance}
\centering
\label{tab: OpenLane 2d cases results}
\scalebox{0.8}{
\begin{tabular}{p{0.25\textwidth}>{\centering}p{0.12\textwidth}>{\centering}p{0.12\textwidth}>{\centering}p{0.12\textwidth}>{\centering}p{0.12\textwidth}>{\centering}p{0.12\textwidth}>{\centering}p{0.13\textwidth}>{\centering\arraybackslash}p{0.13\textwidth}}
\toprule
Method        & All  & \begin{tabular}[c]{@{}c@{}}Up \&\\ Down\end{tabular} & Curve & \begin{tabular}[c]{@{}c@{}}Extreme\\ Weather\end{tabular} & Night & Intersection & \begin{tabular}[c]{@{}c@{}}Merge \&\\ Split\end{tabular} \\ \midrule
LaneATT-S \cite{tabelini2021keep}     & 28.3 & 25.3                                                 & 25.8  & 32.0                                                        & 27.6    &   14.0       & 24.3                                                     \\
LaneATT-M \cite{tabelini2021keep}     & 31.0   & 28.3                                                 & 27.4  &       34.7                                                  & 30.2  &    17.0      & 26.5                                                     \\
CondLaneNet-S \cite{liu2021condlanenet} & 52.3 & 55.3                                                 & 57.5  & 45.8                                                      & 46.6  & 48.4         & 45.5                                                     \\
CondLaneNet-M \cite{liu2021condlanenet} & 55.0   & 58.5                                                 & 59.4  & 49.2                                                      & 48.6  & 50.7         & 47.8                                                     \\
CondLaneNet-L \cite{liu2021condlanenet} & \textbf{59.1} & \textbf{62.1}                                                 & \textbf{62.9}  & \textbf{54.7}                                                      & \textbf{51.0}    & \textbf{55.7}         & \textbf{52.3}                                                     \\
\rowcolor{lightgray}
PersFormer (ours)          & 42.0      &                 40.7                                     &   46.3    &                     43.7                                      &    36.1   &      28.9       &         41.2                                                 \\ \bottomrule
\end{tabular}}
\end{minipage}

\bigskip

\begin{minipage}[t]{\textwidth}

\caption{Comprehensive 3D Lane evaluation under different metrics. On the strength of unified anchor design, PersFormer outperforms previous 3D methods on the metrics of far error while retains comparable results on near error ($m$). $^*$ denotes projecting 2D lane results from CondLaneNet~\cite{liu2021condlanenet} to BEV using IPM}
\centering
\label{tab: OpenLane overall results}
\scalebox{0.8}{
\begin{tabular}{p{0.25\textwidth}>{\centering}p{0.12\textwidth}>{\centering}p{0.14\textwidth}>{\centering}p{0.15\textwidth}>{\centering}p{0.15\textwidth}>{\centering}p{0.15\textwidth}>{\centering\arraybackslash}p{0.15\textwidth}}
\toprule
Method      & F-Score & \begin{tabular}[c]{@{}c@{}}Category\\ Accuracy\end{tabular} & X error near & X error far & Z error near & Z error far \\ \midrule
3D-LaneNet \cite{Garnett_2019_ICCV}  & 44.1    & -                                                           & \textbf{0.479}        & 0.572       & {0.367}        & {0.443}       \\
Gen-LaneNet \cite{guo2020gen} & 32.3    & -                                                           & 0.591        & 0.684       & 0.411         & 0.521        \\
Cond-IPM$^*$ & 36.6 & - & 0.563 & 1.080 & 0.421 & 0.892 \\

\rowcolor{lightgray}
PersFormer (ours)        &     \textbf{50.5}    &           \textbf{92.3}                                                  &       0.485       &      \textbf{0.553}       &      \textbf{0.364}        &       \textbf{0.431}      \\ \bottomrule
\end{tabular}
}

\end{minipage}

\end{table*}


\subsection{Results on OpenLane}
\begin{figure}[t!]
    \centering
    \includegraphics[width=\textwidth]{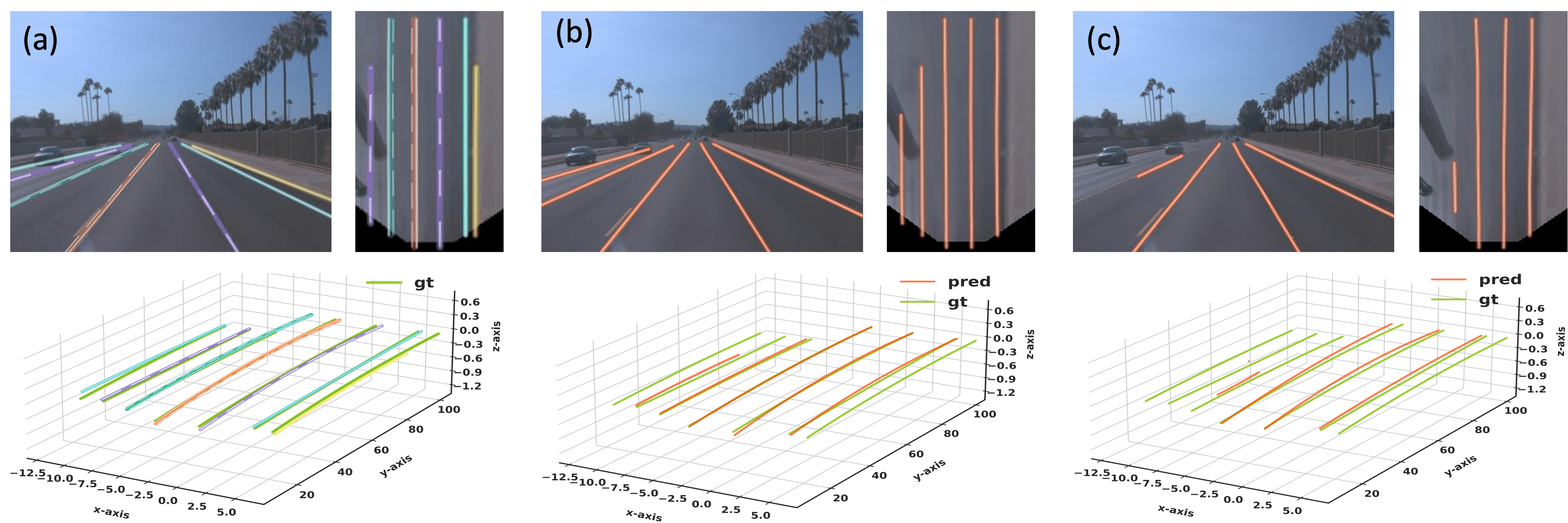}
    \caption{Qualitative results of PersFormer(a), 3D-LaneNet(b) \cite{Garnett_2019_ICCV}, and Gen-LaneNet(c) \cite{guo2020gen}. Under a straight road scenario, PersFormer can provide lane-type information and even detect subtle curbside while other methods are missing it}
    \label{fig:qualitative}
\end{figure}

We provide 3D and 2D evaluation results on the proposed OpenLane dataset. 
%
%
In order to evaluate the models thoroughly, we report F-Score on the entire validation set and different scenario sets.
The scenario sets are selected from the entire validation set based on the scene tags of each frame.
%
%
%
In Tab. \ref{tab: OpenLane 3d cases results}, PersFormer gets the highest F-Score on the entire validation set and every scenario set, surpassing previous SOTA methods in varying degrees.
In Tab. \ref{tab: OpenLane 2d cases results}, PersFormer outperforms LaneATT \cite{tabelini2021keep}, which is our baseline 2D method, by \textbf{11\%}.
%
%
%
%
%
Detailed comparison with previous 3D SOTAs is presented in Tab. \ref{tab: OpenLane overall results}.
PersFormer outperforms the previous best method in F-Score by \textbf{6.4\%}, realizes satisfying accuracy on the classification of lane type, and presents the first baseline result.
Note that PersFormer is not satisfying on the metric of near error on $x$-axis. This is probably because the unified anchor design is more suitable in fitting the main body of a lane rather than the starting point.
Qualitative results are shown in Fig. \ref{fig:qualitative},
indicating that PersFormer is good at catching dense and unapparent lanes in usual autonomous driving scenes.
Overall, PersFormer reaches the best performance on 3D lane detection and gains remarkable improvement in 2D on OpenLane.

\subsection{Results on Apollo 3D Synthetic}
We evaluate PersFormer on Apollo 3D Lane Synthetic dataset \cite{guo2020gen}.
%
%
%
In Tab. \ref{tab: apollot results}, while limited by the scale of the dataset (10K frames), our PersFormer still achieves the best F-Score on every scene set.
In terms of X/Z error, our model gets comparable results compared to previous methods.
%
%


\begin{table}[t!]
\caption{Comparison with previous 3D methods on Apollo 3D Lane Synthetic. PersFormer achieves best F-Score on every scene set with comparable X/Z error ($m$) 
}
\centering
\label{tab: apollot results}
\scalebox{0.8}{
\begin{tabular}{p{0.12\textwidth}p{0.28\textwidth}>{\centering}p{0.12\textwidth}>{\centering}p{0.15\textwidth}>{\centering}p{0.15\textwidth}>{\centering}p{0.15\textwidth}>{\centering\arraybackslash}p{0.15\textwidth}}
\toprule
Scene                                                                      & Method             & F-Score & X error near & X error far & Z error near & Z error far \\ \midrule
\multirow{6}{*}{\begin{tabular}[c]{@{}l@{}}Balanced\\ Scenes\end{tabular}} & 3D-LaneNet \cite{Garnett_2019_ICCV}         & 86.4    & 0.068        & 0.477       & 0.015        & \textbf{0.202}       \\
                                                                           & Gen-LaneNet \cite{guo2020gen}        & 88.1    & 0.061        & 0.496       & 0.012        & 0.214       \\
                                                                           & 3D-LaneNet(l/att) \cite{jin2021robust}  & 91.0      & 0.082        & 0.439       & 0.011        & 0.242       \\
                                                                           & Gen-LaneNet(l/att) \cite{jin2021robust} & 90.3    & 0.080         & 0.473       & 0.011        & 0.247       \\
                                                                           & CLGo \cite{liu2022learning}               & 91.9    & 0.061        & 0.361       & 0.029        & 0.250        \\
                                                                           & \cellcolor{lightgray}PersFormer (ours)               & \cellcolor{lightgray}  \textbf{92.9}    & \cellcolor{lightgray}      \textbf{0.054}       &     \textbf{0.356} \cellcolor{lightgray}            & \cellcolor{lightgray}     \textbf{0.010}        & \cellcolor{lightgray}       0.234     \\ \midrule
\multirow{6}{*}{\begin{tabular}[c]{@{}l@{}}Rarely\\ Observed\end{tabular}} & 3D-LaneNet \cite{Garnett_2019_ICCV}         & 72.0      & 0.166        & 0.855       & 0.039        & \textbf{0.521}       \\
                                                                           & Gen-LaneNet \cite{guo2020gen}        & 78.0      & 0.139        & 0.903       & 0.030         & 0.539       \\
                                                                           & 3D-LaneNet(l/att) \cite{jin2021robust}  & 84.1    & 0.289        & 0.925       & {0.025}        & 0.625       \\
                                                                           & Gen-LaneNet(l/att) \cite{jin2021robust} & 81.7    & 0.283        & 0.915       & 0.028        & 0.653       \\
                                                                           & CLGo \cite{liu2022learning}               & 86.1    & 0.147        & \textbf{0.735}       & 0.071        & 0.609       \\
                                                                           & \cellcolor{lightgray}PersFormer (ours)               & \cellcolor{lightgray}   \textbf{87.5}     & \cellcolor{lightgray}     \textbf{0.107}        & \cellcolor{lightgray}    0.782        & \cellcolor{lightgray}    \textbf{0.024}         & \cellcolor{lightgray}    0.602        \\ \midrule
\multirow{6}{*}{\begin{tabular}[c]{@{}l@{}}Vivual\\ Variants\end{tabular}} & 3D-LaneNet \cite{Garnett_2019_ICCV}         & 72.5    & 0.115        & 0.601       & 0.032        & \textbf{0.230}        \\
                                                                           & Gen-LaneNet \cite{guo2020gen}        & 85.3    & 0.074        & 0.538       & 0.015        & 0.232       \\
                                                                           & 3D-LaneNet(l/att) \cite{jin2021robust}  & 85.4    & 0.118        & 0.559       & 0.018        & 0.290        \\
                                                                           & Gen-LaneNet(l/att) \cite{jin2021robust} & 86.8    & 0.104        & 0.544       & 0.016        & 0.294       \\
                                                                           & CLGo \cite{liu2022learning}               & 87.3    & 0.084        & 0.464       & 0.045        & 0.312       \\
                                                                           & \cellcolor{lightgray}PersFormer (ours)               & \cellcolor{lightgray}    \textbf{89.6}    & \cellcolor{lightgray} \textbf{0.074}      & \cellcolor{lightgray}      \textbf{0.430}      & \cellcolor{lightgray}      \textbf{0.015}       & \cellcolor{lightgray}      0.266      \\ \bottomrule
\end{tabular}}
\end{table}

\subsection{Ablation Study}
We present ablation studies on the anchor design, multi-task strategy, transformer-based view transformation, and auxiliary segmentation task.
We mainly report the improvement on 3D lane detection and provide related results on 2D task.

\textbf{Anchor design and multi-task.}
Starting with a pure 3D lane detection framework (similar to 3D-LaneNet \cite{Garnett_2019_ICCV}), PersFormer gains \textbf{1.7\%} by adopting multi-task scheme (Exp.2) and \textbf{0.98\%} with new anchor design (Exp.4) respectively.
%
%
By jointly using the new anchor and multi-task trick, PersFormer acquires an improvement of \textbf{2.5\%} in 3D task and \textbf{2.6\%} in 2D task (Exp.5).

\textbf{Spatial feature transformation.}
%
By using Perspective Transformer with the new anchor design, the improvement increases to \textbf{4.9\%} (Exp.6), almost doubling the previous improvement.
%
%
Adding auxiliary binary segmentation task further brings an improvement to \textbf{6.02\%} (Exp.7), which is our complete model.
These ablations support our assumption that PersFormer indeed generates a fine-grained BEV feature, and the spatial feature transformation does illustrate its importance in 3D lane detection task.
Surprisingly, a better BEV feature helps 2D task a lot as well, improving \textbf{9.7\%} (Exp.7).
%

\aboverulesep=0ex
\belowrulesep=0ex
\begin{table}[t]
\caption{Ablative Study on a 300 segments subset of OpenLane. Exp.1 is the baseline 3D method, growing with anchor design and multi-task learning (Exp.2-5). The performance culminates with our spatial feature transformation module and explicit BEV supervision (Exp.6,7)}
\centering
\label{tab: ablation}
\scalebox{0.8}{
\begin{tabular}{>{\centering}p{0.06\textwidth}|>{\centering}p{0.1\textwidth}>{\centering}p{0.1\textwidth}>{\centering}p{0.1\textwidth}>{\centering}p{0.13\textwidth}>{\centering}p{0.13\textwidth}|>{\centering}p{0.15\textwidth}>{\centering\arraybackslash}p{0.15\textwidth}}
\toprule
Exp. & \begin{tabular}[c]{@{}c@{}}Unified\\ Anchor\end{tabular} & \begin{tabular}[c]{@{}c@{}}3D\\ Det\end{tabular}  & \begin{tabular}[c]{@{}c@{}}2D\\ Det\end{tabular}  & \begin{tabular}[c]{@{}c@{}}Perspective\\ Transformer\end{tabular} & \begin{tabular}[c]{@{}c@{}}Binary\\ Seg\end{tabular} & 3D F-Score & 2D F-Score \\ \toprule
1    &                                                      & \ding{51} &     &                                                                   &                                                      & 41.77      & -          \\
2    &                                                      & \ding{51} & \ding{51} &                                                                   &                                                      & 43.49      & 32.33            \\
3    & \ding{51}                                                  &     & \ding{51} &                                                                   &                                                      & -          & 34.90       \\
4    & \ding{51}                                                  & \ding{51} &     &                                                                   &                                                      & 42.75      & -          \\
5    & \ding{51}                                                  & \ding{51} & \ding{51} &                                                                   &                                                      & 44.29      & 34.98      \\
6    & \ding{51}                                                  & \ding{51} & \ding{51} & \ding{51}                                                               &                                                      & 46.62      &      37.00      \\
7    & \ding{51}                                                  & \ding{51} & \ding{51} & \ding{51}                                                               & \ding{51}                                                  & 47.79      &      42.00      \\ \bottomrule
\end{tabular}}
\end{table}

\section{Conclusions}
In this paper, we have proposed Persformer, a novel Transformer-based 2D/3D lane detector, along with OpenLane, a large-scale realistic 3D lane dataset. 
We demonstrate experimentally that a fine-grained BEV feature with explicit prior and supervision can significantly improve the performance of lane detection.
Meanwhile, a large-scale real-world 3D lane dataset effectively align the demand from both the academic and the industrial side.
%



\section*{Acknowledgments}
The project is partially supported by the Shanghai Committee of Science and Technology (Grant No. 21DZ1100100). This work was supported in part by National Key Research and Development Program of China (2020AAA0107600), Shanghai Municipal Science and Technology Major Project (2021SHZDZX0102). We would like to acknowledge the great support from SenseBee labelling team at SenseTime Research, constructive contribution from \href{https://github.com/zihanding819}{Zihan Ding at BUAA}, and the fruitful discussions and comments for this project from Zhiqi Li, Yuenan Hou, Yu Liu, Jing Shao, Jifeng Dai.

\clearpage
%
%
\bibliographystyle{splncs04}
\bibliography{egbib}

\clearpage
\appendix
\noindent{\Large \textbf{Appendix}}
\renewcommand{\appendixname}{Appendix~\Alph{section}}
\section{More Related Work}\label{sec: sup}


\subsection{Lane Detection Benchmarks}
For example, \cite{lee2017vpgnet,huang2019apolloscape,yu2020bdd100k} annotate lanes and lane markings in pixel-level so they are best suitable for semantic segmentation task. 
\cite{tusimple2017,llamas2019} collect data on highways with light traffic only, which is not challenging and has a large gap between the evaluation and real-world performance for up-to-date algorithms. 
\cite{pan2018spatial,xu2020curvelane} consider more scenarios under different weather and traffic conditions; however, no-segment character limits their applicability for future applications, such as lane tracking or temporal lane detection. 
The recently released VIL-1000 \cite{zhang2021vil} is specifically designed for video instance lane detection, and yet it does not provide tracking ID annotation across the segments. 
At the same time of our proposing OpenLane dataset, there's another large-scale realistic 3D lane dataset, named ONCE-3DLanes \cite{yan2022once3dlane}, that annotates lane layout in 3D space.
The difference between OpenLane and ONCE-3DLanes falls into three aspects.
First is the dataset statistics. 
The number of frames contained is quite the same, where OpenLane has 200K in total and ONCE-3DLanes has 211K.
The annotation quality differentiates a lot, as OpenLane has more than 25\% of frames with more than 6 lanes, while ONCE-3DLanes only has less than 10\% of frames under the same setting.
Second is the problem setting.
OpenLane provides camera extrinsics as Waymo Open Dataset, while ONCE-3DLanes lacks of this information.
Meanwhile, OpenLane provides segments annotation as scene tags, where ONCE-3DLanes doesn't.
This could be used in video task and expand the potential usage of OpenLane.
Third is the diversity of lane annotation.
In OpenLane, the lane annotation not only contains the 3D position of such a lane, but also several attributes and tracking id.
In ONCE-3DLanes, only the 3D position information is provided.
Due to the difficulty of collecting 3D information for lanes, current 3D lane detection algorithms mainly focus on synthetic data \cite{guo2020gen}. It is small-scale and exists the domain gap between simulation and realistic scenarios. 
%


\subsection{2D Lane Detection}\label{sec: sup - related work}




Early lane detection approaches rely on traditional computer vision techniques, such as filtering \cite{aly2008real,li2016road}, clustering \cite{wang2014approach}, \textit{etc}. With the advent of deep learning, CNN-based methods significantly outperform hand-crafted algorithms.
A typical way is to treat lane detection as a semantic segmentation problem \cite{lee2017vpgnet,pan2018spatial,neven2018towards,hou2019learning,abualsaud2021laneaf}. Binary segmentation \cite{neven2018towards} needs post-clustering process for lane instance discrimination, while multi-class segmentation \cite{lee2017vpgnet,pan2018spatial,hou2019learning} usually limits the maximum detection results in one frame. Moreover, the pixel-wise classification takes large computation resources.
To overcome this, several work propose lightweight yet effective grid based \cite{qin2020ultra,liu2021condlanenet,jayasinghe2021swiftlane,qu2021focus} or anchor based \cite{chen2019pointlanenet,li2019line,xu2020curvelane,su2021structure,tabelini2021keep} methods.
The grid-based approach detects lanes in a row-wise way, whose resolution is much lower than the segmentation map. The model outputs the probability for each cell if it belongs to a lane, and a vertical post-clustering process is still needed to generate the lane instances.
Anchor-based approaches adopt the idea from classical object detection, focusing on optimizing the offsets from predefined line anchors. 
In this circumstance, how to define anchors is a critical problem. Chen \textit{et al.} \cite{chen2019pointlanenet} adopts vertical anchors, which cause great difficulty for curving lane prediction. 
Some work \cite{li2019line,tabelini2021keep,su2021structure} design anchors as 
a slender tilt shape, while the huge amount of different anchors to improve the detection accuracy would influence the computational efficiency. 
Nevertheless, considering their incredible performance on public datasets, we adopt the anchor-based formulation and carefully re-design anchors to achieve both high accuracy and efficiency.

\section{Algorithm}\label{sec: sup - alg}
We summarize the details of PersFormer here. We introduce the backbone, overall structure and the unified anchor design. Later we break down the loss function into pieces.
\subsection{Backbone}
%
%
The backbone module is slightly different from previous work \cite{Garnett_2019_ICCV,guo2020gen}, as we need to consider 2D/3D branches together.
We use EfficientNet \cite{tan2019efficientnet} as our backbone, and extract a specific layer as our following module's input. 
%
%
%
Later we provide two designs, using FPN \cite{lin2017fpn} or not.
After using several convolution layers, the backbone module outputs 4 different scaled front-view feature maps.
Their resolutions are $180 \times 240$, $90 \times 120$, $45 \times 60$, $22 \times 30$.
Each front-view feature map is then transformed to BEV-space feature map with the help of Perspective Transformer, resulting in 4 BEV feature maps.
%

\subsection{Anchor Details}

In this section, we present details of our anchor design, including angles, numbers of anchors and how we associate ground truth lanes with anchors in 2D and 3D.
As introduced in the main body of the paper, we first set anchors in BEV space.
Following Gen-LaneNet \cite{guo2020gen}, the starting positions $X^i_{\text{bev}}$ are evenly placed along $x$-axis with the spacing of 8 pixels.
However, we differentiate it from the incline angle $\varphi$. Gen-LaneNet sets straight-forward (parallel to $y$-axis) only, which makes it hard to predict lanes with large curvatures or perpendicular lanes.
Towards this problem, we put 7 anchors at each $X^i_{\text{bev}}$ with different angles, \textit{i.e.}, $\varphi \in \{\pi/2, \arctan{(\pm 0.5)}, \arctan{(\pm 1)}, \arctan{(\pm 2)} \}$. Note that the angles are in terms of grid coordinates, which is not equal to the absolute values when grids are not square.
Moreover, we project all the BEV anchors to image space with average camera height and pitch angle of the dataset, leading to corresponding 2D anchors.

The association between ground truth lanes and anchors is based on the average distance similar to the loss calculation process, instead of assigning the closest anchor at $Y_{ref}$ to ground truths as \cite{Garnett_2019_ICCV,guo2020gen}.
The $Y_{ref}$ is set very close to ego-vehicle, \textit{i.e.}, $5m$ in Gen-LaneNet, which makes it better predict lanes in close area while having unsatisfactory performance in the far distance.
In our experiments, we assign the anchor with minimum \textit{edit distance} to ground truth lanes in both 2D and 3D tasks.
The distance is calculated at fixed $y$ positions: $(5, 10, 15, 20, 30, 40, 50, 60, 80, 100)$ for 3D anchors, and 72 equally sampled heights for 2D anchors.

\subsection{Loss Function}
We give the details of loss function here.
As introduced in the main body of the paper, given the pre-defined \textit{y} value of the $N_d$ samples along \textit{y}-axis, the 3D detection head outputs a set of points for each anchor $i$ as following:
\begin{equation}
    (\mathbf{x}^i, \mathbf{z}^i, \mathbf{vis}^i_{\text{bev}}) = \{ (x^{(i,k)}, z^{(i,k)}, \text{vis}^{(i,k)}_{\text{bev}} ) \}^{N_d}_{k=1}
\end{equation}
%
The \textit{y} values are $(5, 10, 15, 20, 30, 40, 50, 60, 80, 100)$ in the BEV space, and the size of the BEV space is $20m \times 100m$.
Similar to 3D setting, given the pre-defined \textit{v} value of the $N_d$ samples along \textit{v}-axis in front view, the 2D prediction is:
\begin{equation}
    (\mathbf{u}^i, \mathbf{vis}^i_{\text{uv}}) = \{ (u^{(i,k)}, \text{vis}^{(i,k)}_{\text{uv}} ) \}^{N_d}_{k=1}
\end{equation}
The loss is a combination of the 2D lane detection, 3D lane detection and intermediate segmentation with learnable weights $(\alpha, \beta, \gamma)$ accordingly:
\begin{equation}
        \mathcal{L} = 
        \sum_i  
        \alpha \mathcal{L}_{ \text{2D}  }( c^i_{  \text{2D}   }, \mathbf{u}^i, \mathbf{vis}^i_{\text{fv}} ) + \beta 
        \mathcal{L}_{ \text{  3D } } ( c^i_{ \text{3D} }, \mathbf{x}^i, \mathbf{z}^i, \mathbf{vis}^i_{\text{bev}} ) + \gamma \mathcal{L}_{ \text{seg}} (S_{\text{pred}}),
\end{equation}
where $c^i_{( \cdot )} $ is the predicted lane category in 2D and 3D domain respectively.
For $\mathcal{L}_{ \text{  3D } }$, it consists of classification loss, regression loss and visibility loss.
The classification loss is a cross-entropy loss, which is as follow:
\begin{equation}
    \mathcal{L}_{\text{3D}{\text -}\text{cls}} = \mathcal{L}_{CE}(c^i_{\text{3D}{\text -}\text{pred}}, c^i_{\text{3D}{\text -}\text{gt}})
\end{equation}
The regression loss is a $L_1$ loss, which is as follow:
\begin{equation}
    \mathcal{L}_{\text{3D}{\text -}\text{reg}} = \mathcal{L}_{L_1}(\{\mathbf{x}^i, \mathbf{z}^i\}_{\text{pred}}, \{\mathbf{x}^i, \mathbf{z}^i\}_{\text{gt}})
\end{equation}
The visibility loss is a binary cross-entropy loss, which is as follow:
\begin{equation}
    \mathcal{L}_{\text{3D}{\text -}\text{vis}} = \mathcal{L}_{BCE}(\mathbf{vis}^i_{\text{pred}}, \mathbf{vis}^i_{\text{gt}})
\end{equation}
The 2D loss functions are similar to the 3D ones, except they are in 2D form:
\begin{equation}
    \begin{split}
        \mathcal{L}_{\text{2D}{\text -}\text{cls}} &= \mathcal{L}_{CE}(c^i_{\text{2D}{\text -}\text{pred}}, c^i_{\text{2D}{\text -}\text{gt}}) \\
        \mathcal{L}_{\text{2D}{\text -}\text{reg}} &= \mathcal{L}_{L_1}(\{\mathbf{u}^i \}_{\text{pred}}, \{\mathbf{u}^i\}_{\text{gt}}) \\
        \mathcal{L}_{\text{2D}{\text -}\text{vis}} & = \mathcal{L}_{BCE}(\mathbf{vis}^i_{\text{pred}}, \mathbf{vis}^i_{\text{gt}})
    \end{split}
\end{equation}
The segmentation loss is a binary cross-entropy loss as well, which is as follow:
\begin{equation}
    \mathcal{L}_{\text{seg}} = \mathcal{L}_{BCE}(S_{\text{pred}}, S_{\text{gt}})
\end{equation}


\section{Details on OpenLane Benchmark}\label{sec: sup - dataset}

In this section, we present more details on dataset statics, our annotation criterion, visualization examples, algorithms we adopted when generating the dataset.

\subsection{Dataset Statistics}

OpenLane has 1,150 segments with train/validation/test splits of 798/202/150, respectively. Since the test sets are kept for its online leaderboard evaluation, we annotate the other 1,000 segments, \textit{i.e.}, 200K frames at a frequency of 10 FPS, and keep the original train/validation partition for fair comparison 
with other tasks, such as object detection.

\begin{figure}[t!]
    \centering
      \includegraphics[width=.9\textwidth]{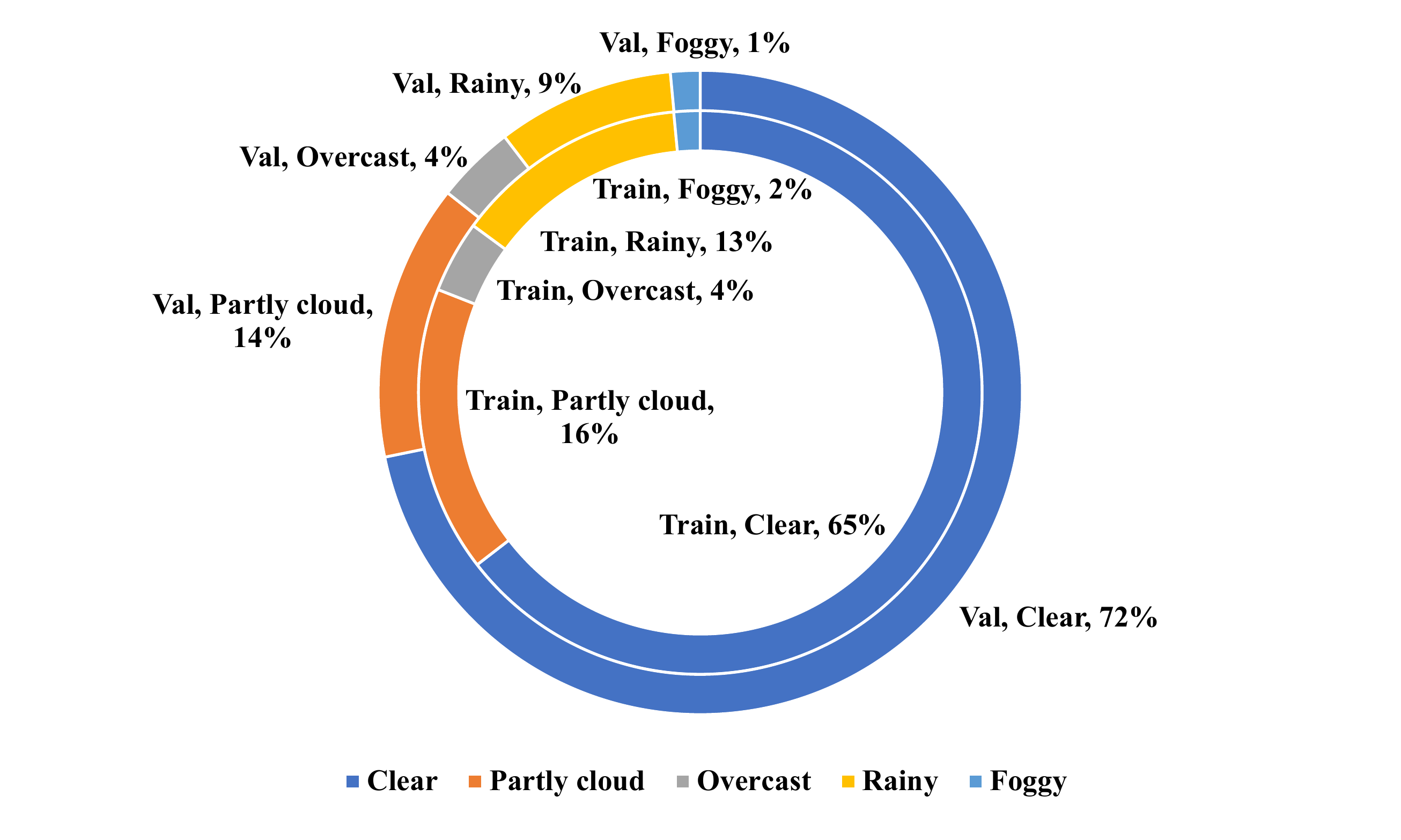}
    \vspace{-.2cm}
      \caption{Distribution of weather tags in training and validation sets. The data is collected under different weathers and split into training and validation with great balance}
    \label{fig: stat - weather}
\end{figure}

\begin{figure}[t!]
    \centering
      \includegraphics[width=.9\textwidth]{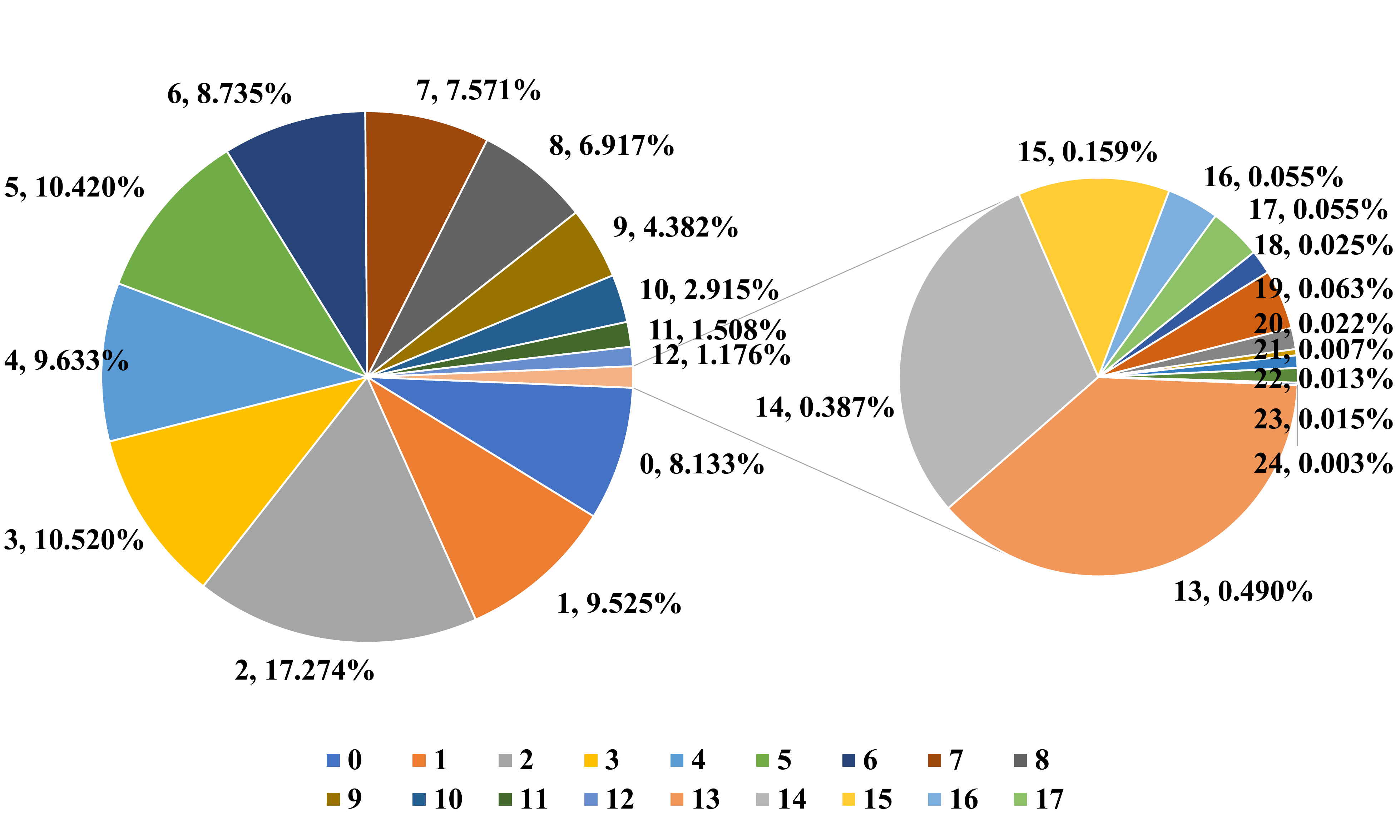}
    \vspace{-.2cm}
      \caption{Distribution of lane numbers per frame. The maximum number is 24, and 25\% frames have more than 6 lane}
    \label{fig: stat - number}
\end{figure}

\begin{figure}[t!]
    \centering
      \includegraphics[width=.9\textwidth]{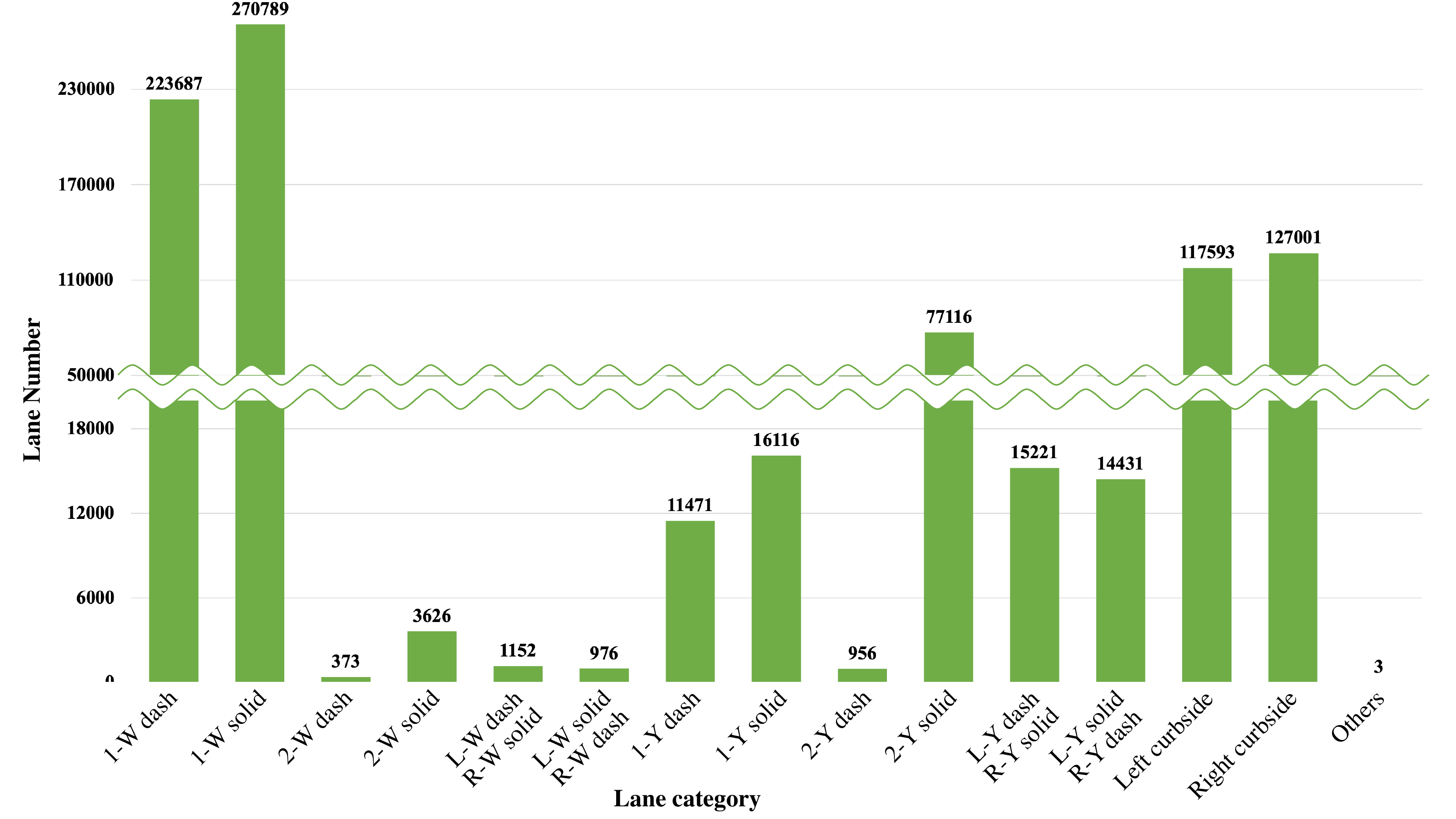}
    \vspace{-.2cm}
      \caption{Distribution of the lane category. Here we abbreviate single in \textit{1}, double in \textit{2}, white in \textit{W}, yellow in \textit{Y}, left in \textit{L}, and right in \textit{R}. Thus \textit{1-W dash} means the category of single white dash lanes}
    \label{fig: stat - type}
\end{figure}

\begin{figure}[t!]
    \centering
      \includegraphics[width=.9\textwidth]{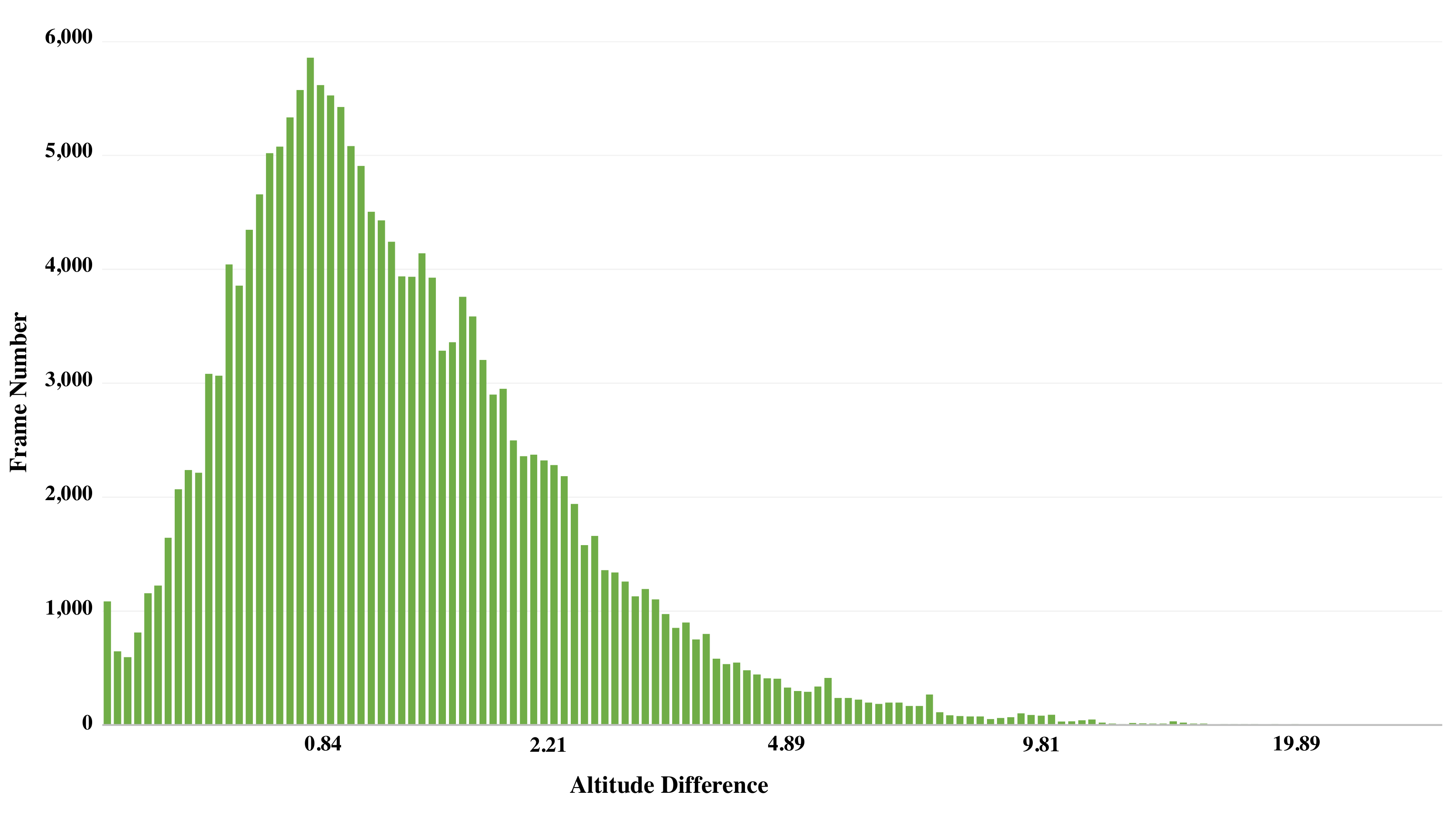}
    \vspace{-.2cm}
      \caption{Altitude difference per frame. Note the x-axis is approximately in a log scale and its unit is $m$}
    \label{fig: stat - height}
\end{figure}

We compute the statistics in OpenLane and visualize them.
The overall number of segments with different scene tags is given in Tab. \ref{tab: stat - tags}. It implies great diversity in data collection and raises higher requirements on the robustness of algorithms.
The weather distribution is visually presented 
in Fig. \ref{fig: stat - weather}. It shows the benchmark 
covers various weather conditions and well holds the consistency in the train/validation split.
The distribution of the number of lanes in each frame is shown in Fig. \ref{fig: stat - number}. About 25\% frames of OpenLane have more than 6 lanes, which exceeds the maximum number in most lane datasets. 
Fig. \ref{fig: stat - type} shows the distribution of lane categories. 
Single white solid and dash lanes, double yellow solid lanes take up almost 90\% of the total lanes. This is imbalanced and yet it falls into a long-tail distribution problem, which is common in realistic scenarios. 
Fig. \ref{fig: stat - height} presents the distribution of altitude difference per frame. Only around 20\% frames are relatively flat with absolute height variation less than 0.5$m$, whereas the difference is more than 1$m$ in over 50\% of OpenLane. This data further demonstrates the necessity of 3D lane detection.
The above statistics and examples below demonstrate that OpenLane is the most challenging one compared to existing lane detection datasets.

\begin{table}[tb!]
\caption{Statics of scenario tags. Scene tags are annotated in terms of segments}
\centering
\label{tab: stat - tags}
\scalebox{0.8}{
\begin{tabular}{p{0.13\textwidth}p{0.15\textwidth}|>{\centering}p{0.1\textwidth}>{\centering}p{0.1\textwidth}>{\centering\arraybackslash}p{0.1\textwidth}}
\toprule
\multicolumn{2}{c}{Tags}                & Train & Val. & All \\ \midrule
\multirow{5}{*}{Weather} & Clear        & 515   & 145  & 660 \\
                         & Partly cloud & 131   & 28   & 159 \\
                         & Overcast     & 33    & 8    & 41  \\
                         & Rainy        & 107   & 18   & 125 \\
                         & Foggy        & 12    & 3    & 15  \\ \midrule
\multirow{5}{*}{Scene}   & Residential  & 270   & 69   & 339 \\
                         & Urban        & 234   & 56   & 290 \\
                         & Suburbs      & 259   & 64   & 323 \\
                         & Highway      & 30    & 6    & 36  \\
                         & Parking lot  & 5     & 7    & 12  \\ \midrule
\multirow{3}{*}{Hours}   & Daytime      & 653   & 167  & 820 \\
                         & Night        & 88    & 22   & 110 \\
                         & Dawn/Dusk    & 57    & 13   & 70  \\ \bottomrule
\end{tabular}}
\end{table}

\subsection{Annotation Criterion}

We aim at introducing how we annotate lanes, scene tags and CIPO levels in this section. Details such as data structures, folder hierarchy will be provided in the dataset releasing page in the future.

\begin{figure}[t!]
    \centering
    \includegraphics[width=\textwidth]{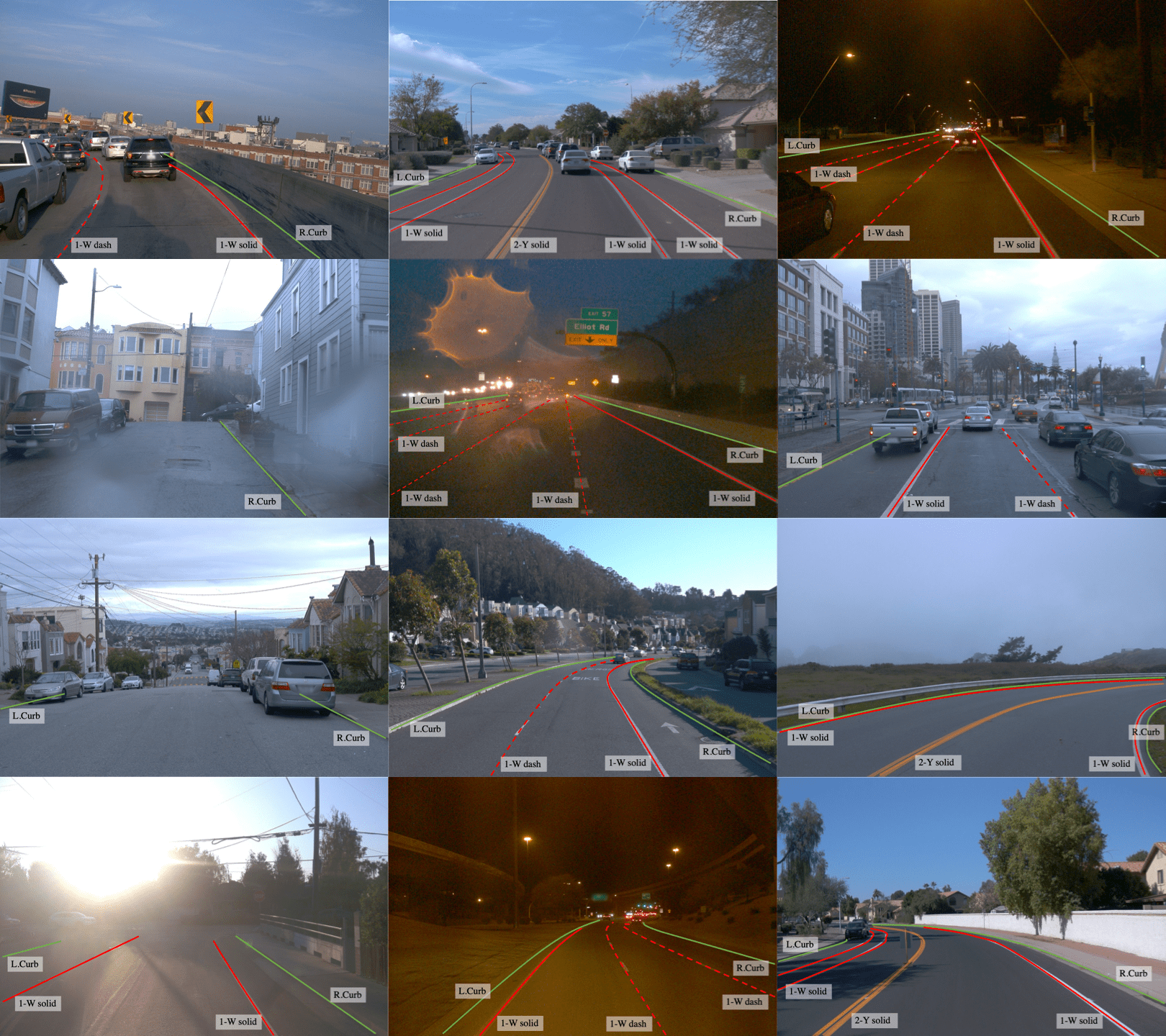}
    \caption{Visualization example of lane annotation in OpenLane dataset}
    \label{fig:sup-dataset example-lane}
\end{figure}

\textbf{Lanes.}
Our principle for the 2D lane detection task is to find all visible lanes inside left and right road edges.
Following this philosophy, we carefully annotate lanes in each frame. However, due to the complexity of scenarios, there exist some special cases we seek to illustrate here.
(1) Lanes are often occluded by objects or invisible because of abrasion but they are still valuable for the real application. Thus we annotate lanes if parts of them are visible, meaning lanes with one side being occluded are extended or lanes with invisible intermediate parts are completed according to the context, as shown in Fig. \ref{fig:sup-dataset example-lane}.
(2) It is very common that the number of lanes changes, especially when lanes have complex topologies such as fork lanes in merge and split cases. Traditional lane datasets usually omit these scenarios for simplicity, while we keep them all and further choose them out of the whole dataset for evaluation. Fork lanes are annotated as separate lanes with a common starting point (split) or ending point (merge) - two close adjacent lanes are desired for the lane detection methods.
(3) We further annotate each lane as one of the 14 lane categories, \textit{i.e.}, single white dash, single white solid, double white dash, double white solid, double white dash solid (left white dash with right white solid), double white solid dash (left white solid with right white dash), single yellow dash, single yellow solid, double yellow dash, double yellow solid, double yellow dash solid (left yellow dash with right yellow solid), double yellow solid dash (left yellow solid with right yellow dash), left curbside, right curbside. Note that traffic bollards are considered as curbsides as well if they are not temporally placed.
(4) Different from all the other lane datasets, we annotate a tracking ID for each lane which is unique across the whole segment. We believe this could be helpful for video lane detection or lane tracking tasks.
We also assign a number in 1-4 to the most important 4 lanes based on their relative position to the ego-vehicle. Basically, the left-left lane is 1, the left lane is 2, the right lane is 3, and the right-right lane is 4.

All valid 2D ground truths are transformed to 3D annotations by the generation method in Sec. \textcolor{red}{4.2} of the main body (Generation of High-quality Annotation), except those without LiDAR points scanning through. Thus the criterion above applies to 3D lanes as well.

\textbf{Scene tags.}
We label each segment with 3 scene tags, \textit{i.e.}, weather, scene and hours. We hope these labels can help researchers to investigate the robustness of their models under various scenarios.
The statics are shown in Tab. \ref{tab: stat - tags}.
Specifically, the dataset covers 5 different kinds of weather, clear, partly cloud, overcast, rainy and foggy. Note that we classify the video as partly cloud or foggy when there are clouds or fog in the sky respectively, otherwise it will be categorized as overcast.
The scene, or the location, includes 5 categories, \textit{i.e.}, residential, urban, suburbs, highway and parking lot.
And the hours are divided into 3 parts: daytime, night, dawn/dusk.

\begin{figure}[t!]
    \centering
    \includegraphics[width=\textwidth]{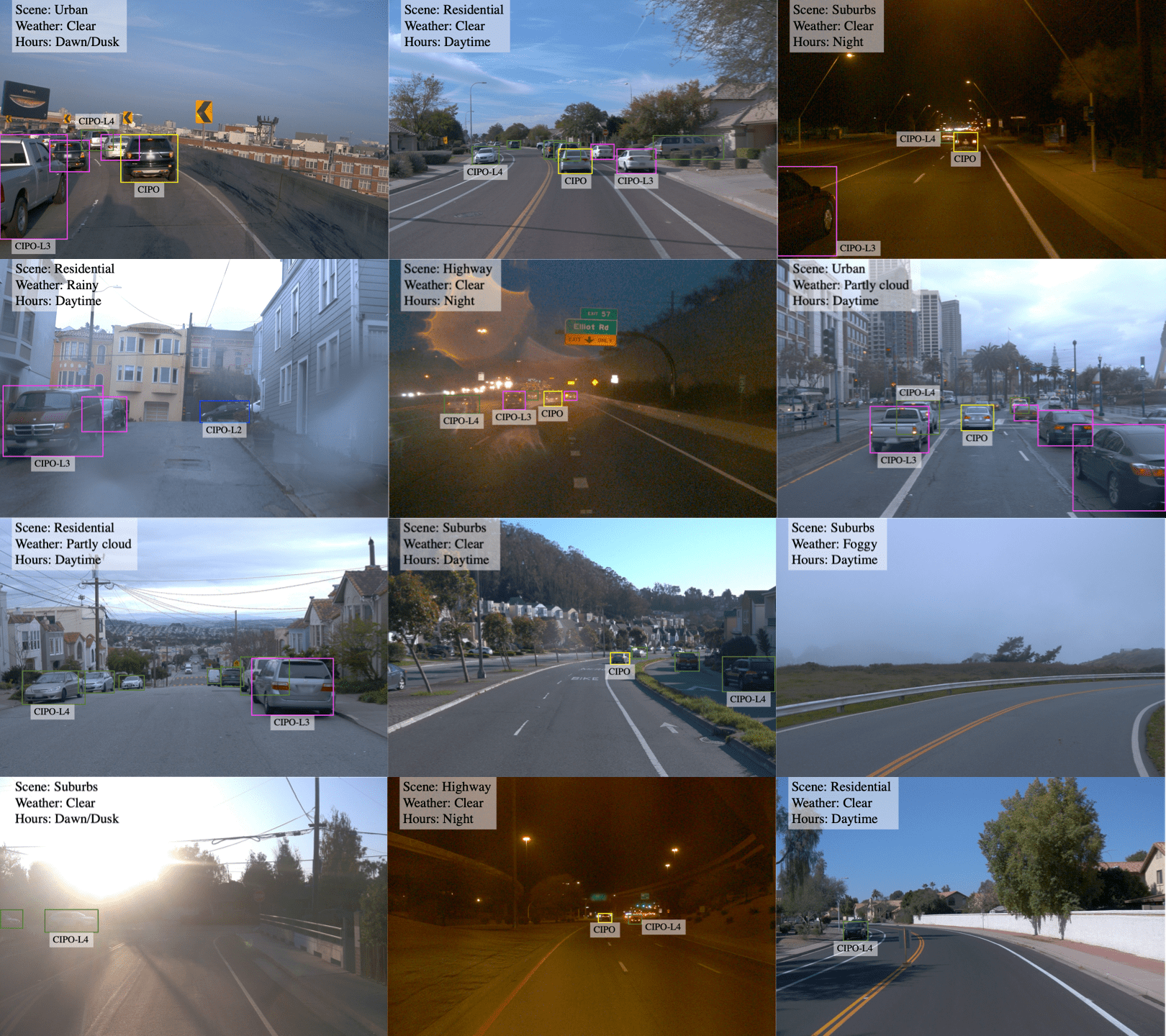}
    \caption{Visualization example of CIPO and Scene tags annotation in OpenLane dataset}
    \label{fig:sup-dataset example-cipo}
    \vspace{-4pt}
\end{figure}

\textbf{Closest-in-path object (CIPO).}
CIPO is usually defined as the closest object in ego lane, which refers to a single vehicle only.
However, there are cases that vehicles on left/right lanes are intended to cut in which are crucial as well, or there may not be any qualified vehicles in ego lane.
To cover the complex scenarios, we categorize objects, mainly including vehicles, pedestrians and cyclists, into 4 different CIPO levels.
(1) The most important one, which is closest to ego vehicle within the required reaction distance and has over 50\% part of it in the ego lane. Level 1 contains one object at most.
(2) Objects are annotated as Level 2 when their bodies interact with the real or virtual lines of ego lane. They are typically in the process of cut-in or cut-out, which hugely influences ego-vehicle decision-making.
(3) We consider objects mainly within the reaction distance or drivable area, or those in left/ego/right lanes more specifically. Thus we annotate Level 3 with objects in the above area and having occlusion rate less than 50\%.
Note that vehicles in the opposite direction can be in this CIPO level as well.
(4) The remainings are labeled as Level 4, which means they are almost unlikely to impact the future path at this moment. They are mainly objects in lanes with far distance, objects out of drivable area, or parked vehicles in our dataset.
Examples are provided in Fig. \ref{fig:sup-dataset example-cipo}.




\subsection{3D lane Generation}

\begin{figure}[tb!]
    \centering
    \includegraphics[width=\textwidth]{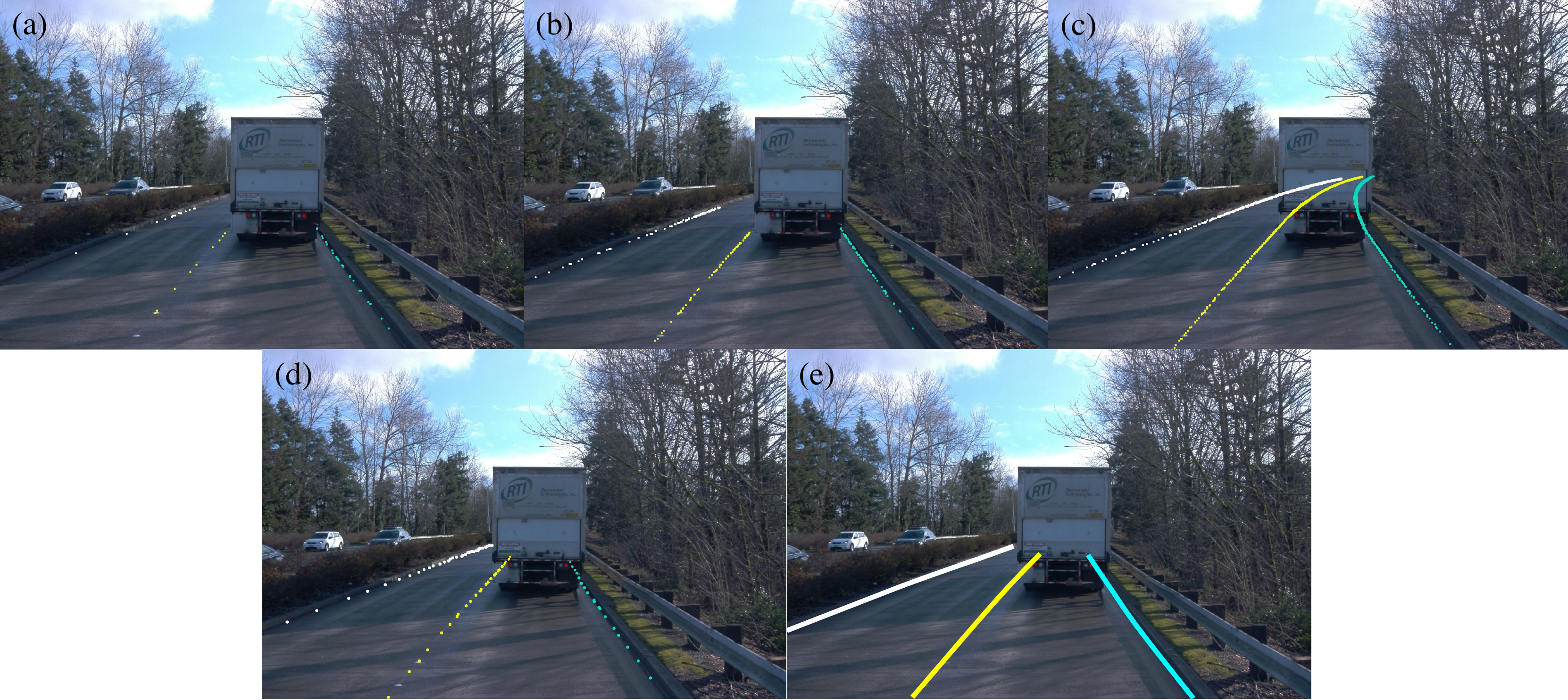}
    \caption{3D lane generation pipeline. (a) Original point clouds inside a certain threshold of 2D lane annotations are reserved, which is relatively sparse; (b) Positions of points on the 2D annotation are interpolated to get a dense point set; (c) 3D lane points in the same segment are spliced into long, high-density lanes; (d) We remove those too far as they are invisible, while reasonable extensions are desired; (e) A smooth and fitting process is applied to get the final 3D lane annotation}
    \label{fig:sup-dataset gen pipeline}
\end{figure}

\begin{figure}[tb!]
    \centering
    \includegraphics[width=\textwidth]{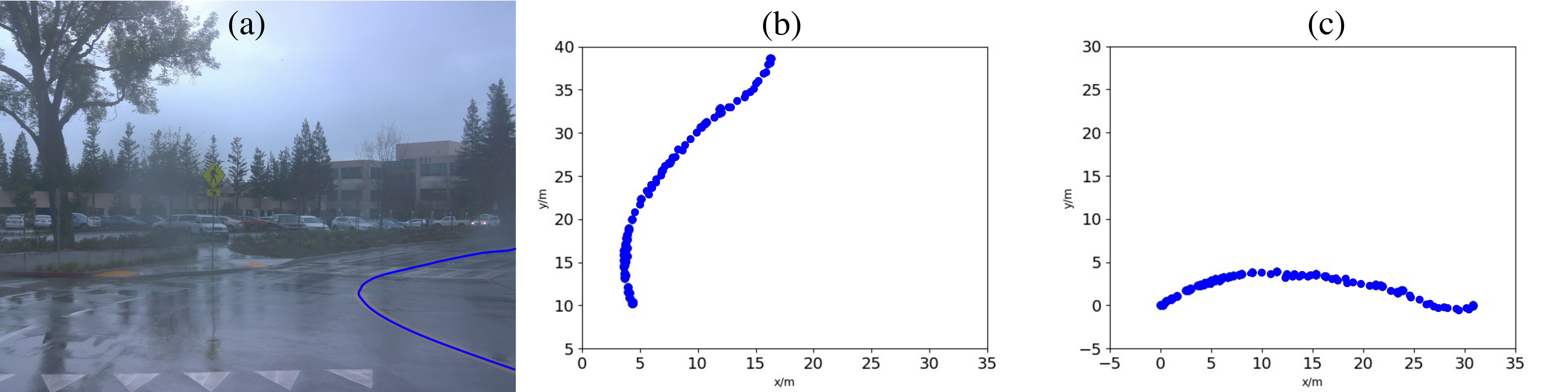}
    \caption{Illustration of 3D lane generation problem with large curvatures. (a) The original image and the 2D lane; (b) Unsorted 3D points set of the lane in (a), which a filtering algorithm is not applicable directly; (c) A simple translation and rotation can result in a one-to-one mapping of $x$ and $y$}
    \label{fig:sup-dataset large curvature}
\end{figure}

Fig. \ref{fig:sup-dataset gen pipeline} shows the intermediate results of the generation process of 3D lane labels.
However, the above process could have a few problems in some cases, especially in the last step, \textit{i.e.}, smoothing and fitting.
Multiple filtering and fitting algorithms are adopted to realize it, while all of them require a set of sorted points.
Due to the large curvature, the one-to-one mapping probably does not stand either in $x$ or $y$ direction, thus we could not sort the points directly. Towards this problem, for each image with this circumstance, we simply find an angle to rotate the whole points set, do the filtering and fitting process in the temporary coordinate and rotate back in the end. This method is illustrated in Fig. \ref{fig:sup-dataset large curvature}.


\section{Experiments}\label{sec: sup - exp}

\subsection{Evaluation Metrics.}
For both 3D lane datasets, we follow the evaluation metric designed by Gen-LaneNet \cite{guo2020gen}, with small modifications\footnote{Please see in OpenLane page: \url{https://github.com/OpenPerceptionX/OpenLane}.} and additional category accuracy on OpenLane dataset. 
The matching between prediction and ground truth is built upon \textit{edit distance}, where one predicted lane is considered to be a true positive only if 75\% of its covered y-positions have a point-wise distance less than the max-allowed distance (1.5$m$).
Then, with the percentage of matched ground-truth lanes as recall and the percentage of matched prediction lanes as precision, we use F-score to report the regression performance of such a model.
Since OpenLane dataset has category information per lane, we present the accuracy upon the matched lanes to show classification performance.
We only report the accuracy of PersFormer on OpenLane dataset, as other 3D methods do not support classification task.
For the 2D task, the classical metric in CULane \cite{pan2018spatial} is adopted.


\subsection{Implementation Details}
To fairly compare with other methods  \cite{guo2020gen,Garnett_2019_ICCV,liu2022learning}, we retain many model settings of image resolution and BEV scale.
We resize the original image to $360\times480$ as model input, project it to BEV space with a resolution of $208 \times 108$. 
%
%
%
We use PyTorch \cite{paszke2019pytorch} to implement the model.
The batch size is set to 8; the number of training epochs is set to 100.
%
We re-implement 3D-LaneNet and Gen-LaneNet on OpenLane dataset for a fair comparison.
%
Following previous experience on training vision transformer  \cite{carion2020end,zhu2021deformable,wang2022detr3d}, we use Adam optimizer \cite{2015KingmaBadam} with base learning rate of $2 \times 10^{-4}$, $\beta_1 = 0.9$, $\beta_2 = 0.999$ and weight decay of $10^{-4}$.
All of these models are trained on 8 NVIDIA Tesla V100 GPUs.
More details about environment setup can be referred to our GitHub repository once accepted.

\subsection{More Experimental Results}
In this section, we present more experimental results, mainly in 3D comparison on ONCE-3DLanes, additional ablations and more qualitative examples.

%
\subsubsection{3D Comparisons on ONCE-3DLanes.}
We provide additional experimental results on ONCE-3DLanes dataset \cite{yan2022once3dlane}, as it's another real-world 3D lane dataset concurrently presented.
ONCE-3DLanes also uses F-Score as the evaluation metric, and more details can be found in their repo \href{https://github.com/once-3dlanes/once_3dlanes_benchmark}{ONCE-3DLanes}.
In Tab.~\ref{tab: ONCE results}, PersFormer gets the highest F-Score on the validation set, outperforming its proposed method SALAD \cite{yan2022once3dlane} over \textbf{10\%}.
One thing worth noticing is that ONCE-3DLanes does not provide camera extrinsics, therefore PersFormer pre-define a set of extrinsic parameters to fit the model setting.
The camera height is set to be 1.5$m$ and pitch to be 0.5.
This does not affect the evaluation results since it is just to fit the IPM process in PersFormer.

\begin{table*}[t!]
\caption{New results on the new benchmark (CVPR22) ONCE-3DLanes~\cite{yan2022once3dlane}. $^*$ denotes results from the paper~\cite{yan2022once3dlane}}
\centering
\label{tab: ONCE results}
\scalebox{0.8}{
\begin{tabular}{p{0.25\textwidth}>{\centering}p{0.15\textwidth}>{\centering}p{0.15\textwidth}>{\centering}p{0.15\textwidth}>{\centering\arraybackslash}p{0.15\textwidth}}
\toprule
Method      & F-Score(\%)  & Precision(\%) & Recall(\%) & CD error($m$) \\ \midrule
3D-LaneNet$^*$~\cite{Garnett_2019_ICCV}  & 44.73  & 61.46 & 35.16 & 0.127 \\
Gen-LaneNet$^*$~\cite{guo2020gen} & 45.59 & 63.95  & 35.42  & 0.121 \\
SALAD$^*$~\cite{yan2022once3dlane} & 64.07 & 75.90 & 55.42 & 0.098 \\  
\rowcolor{lightgray}
\textbf{PersFormer} (ours) & \textbf{74.33}  & \textbf{80.30} & \textbf{69.18} & \textbf{0.074} \\ \bottomrule
\end{tabular}}
\end{table*}

\subsubsection{Ablations.}
We provide an additional ablative study on the structure of the feature transformation module on a subset of OpenLane ($\sim$300 segments) in Tab. \ref{tab: ablation_2}.
%
%
%
We argue that the IPM-based cross attention is a necessity in PersFormer, as we compare it with two initial designs, naive one-to-one mapping and the learned mapping.
The naive one-to-one mapping simply scales every location in the BEV space to the corresponding location in the front view space, not considering camera parameters (Exp.1).
A more ``aggressive" way to simulate the mapping is directly learning from the front view feature with several fully-connected layers (Exp.2).
Neither of them could catch up with the performance of IPM-based mapping, indicating the importance of such a prior in generating BEV feature.
We further attempt to adopt Multi-scale Deformable Attention from \cite{zhu2021deformable} to implement a several-for-one feature mapping from multi-scale front view feature to multi-scale BEV feature (Exp.3), just like Deformable DETR.
The result slightly falls behind our final design (Exp.5), probably due to the influence of tuning of hyper-parameters and the impact of the small-scale feature on the large-scale feature.
Finally, we try to remove the classical self attention module in ordinary Transformer design (Exp.4), showing that the self attention module is all there for a reason in Transformer-style structure.

\begin{table}[tb!]
\caption{Ablative Study on PersFormer Design. IPM prior plays a vital role in guiding the generation of BEV feature compared to naive one-to-one mapping and learned reference-target mapping. Using MSDeformAttn from Deformable DETR \cite{zhu2021deformable} to map multi-scale front-view feature to multi-scale BEV feature is competitive, and the self-attention module of BEV query is important in Transformer-style structure}
\centering
\label{tab: ablation_2}
\scalebox{0.8}{
\begin{tabular}{>{\centering}p{0.06\textwidth}|>{\centering}p{0.1\textwidth}>{\centering}p{0.12\textwidth}>{\centering}p{0.1\textwidth}>{\centering}p{0.1\textwidth}>{\centering}p{0.1\textwidth}|>{\centering\arraybackslash}p{0.15\textwidth}}
\toprule
Exp. & \begin{tabular}[c]{@{}c@{}}Naive\\ 1-1\end{tabular} & Learned  & \begin{tabular}[c]{@{}c@{}}Multi-to-\\ Multi\end{tabular}  & \begin{tabular}[c]{@{}c@{}}Self\\ Attn.\end{tabular} & \begin{tabular}[c]{@{}c@{}}IPM\\ Prior\end{tabular} & 3D F-Score \\ \toprule
1    &                \ding{51}                                      &  &     &                                                                   &                                                      & 36.15          \\
2    &                                                      & \ding{51} &  &                                                                   &                                                      & 13.45               \\
3    &                                                      &  & \ding{51} &                                                                   &                                                      & 51.35               \\
4    &                                                  &  &     &    \ding{51}                                                               &                                                      & 47.18          \\
5    &                                                  & &  &                                                                   &                    \ding{51}                                  & 52.68           \\
\bottomrule
\end{tabular}}
\end{table}

\subsubsection{Visualization.}
\begin{figure}[tb!]
    \centering
    \includegraphics[width=\textwidth]{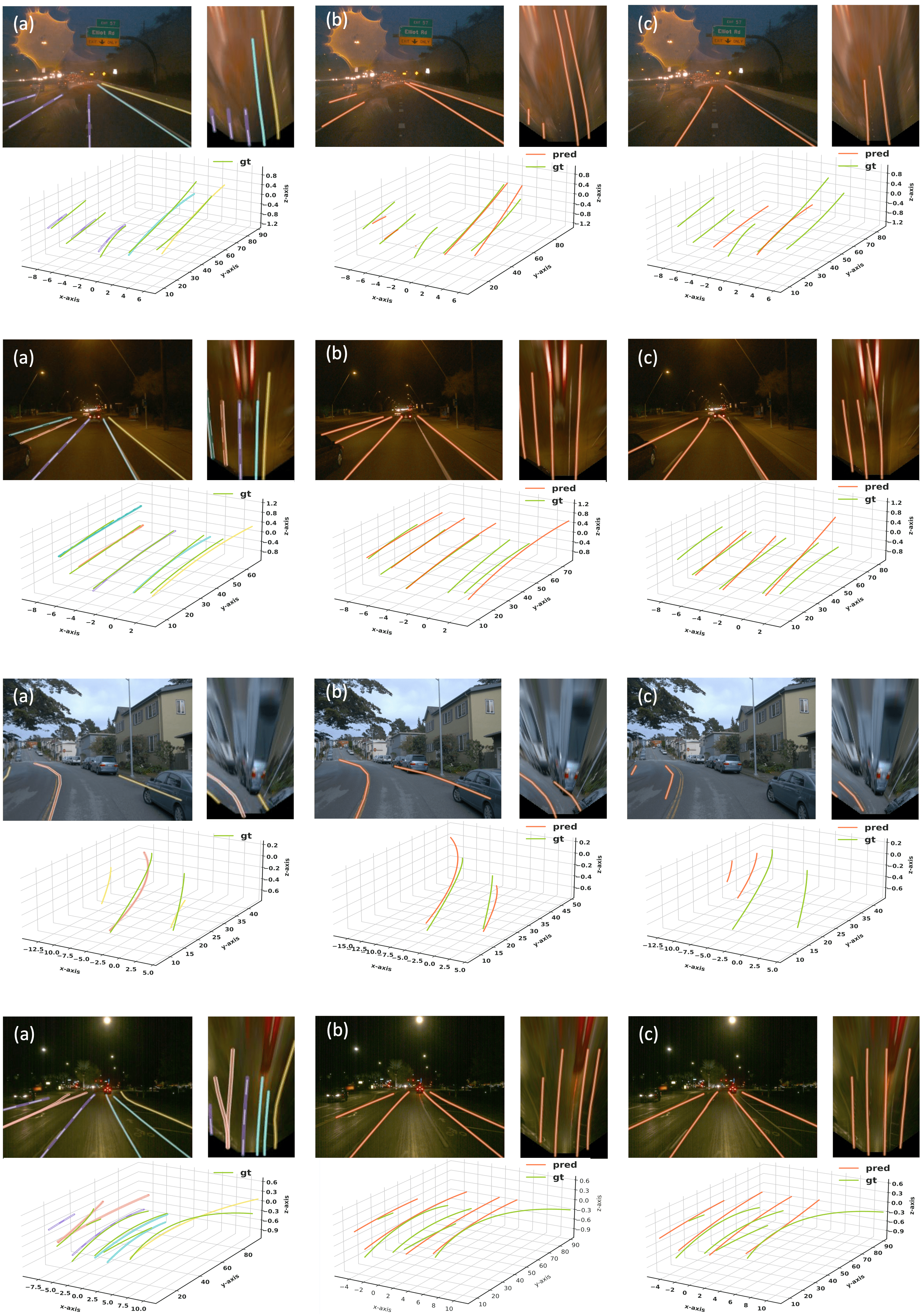}
    \caption{Qualitative results of PersFormer(a), 3D-LaneNet(b) \cite{Garnett_2019_ICCV}, and Gen-LaneNet(c) \cite{guo2020gen} on OpenLane. Night case and Up\&Down case}
    \label{fig:qualitative-1}
\end{figure}

\begin{figure}[tb!]
    \centering
    \includegraphics[width=\textwidth]{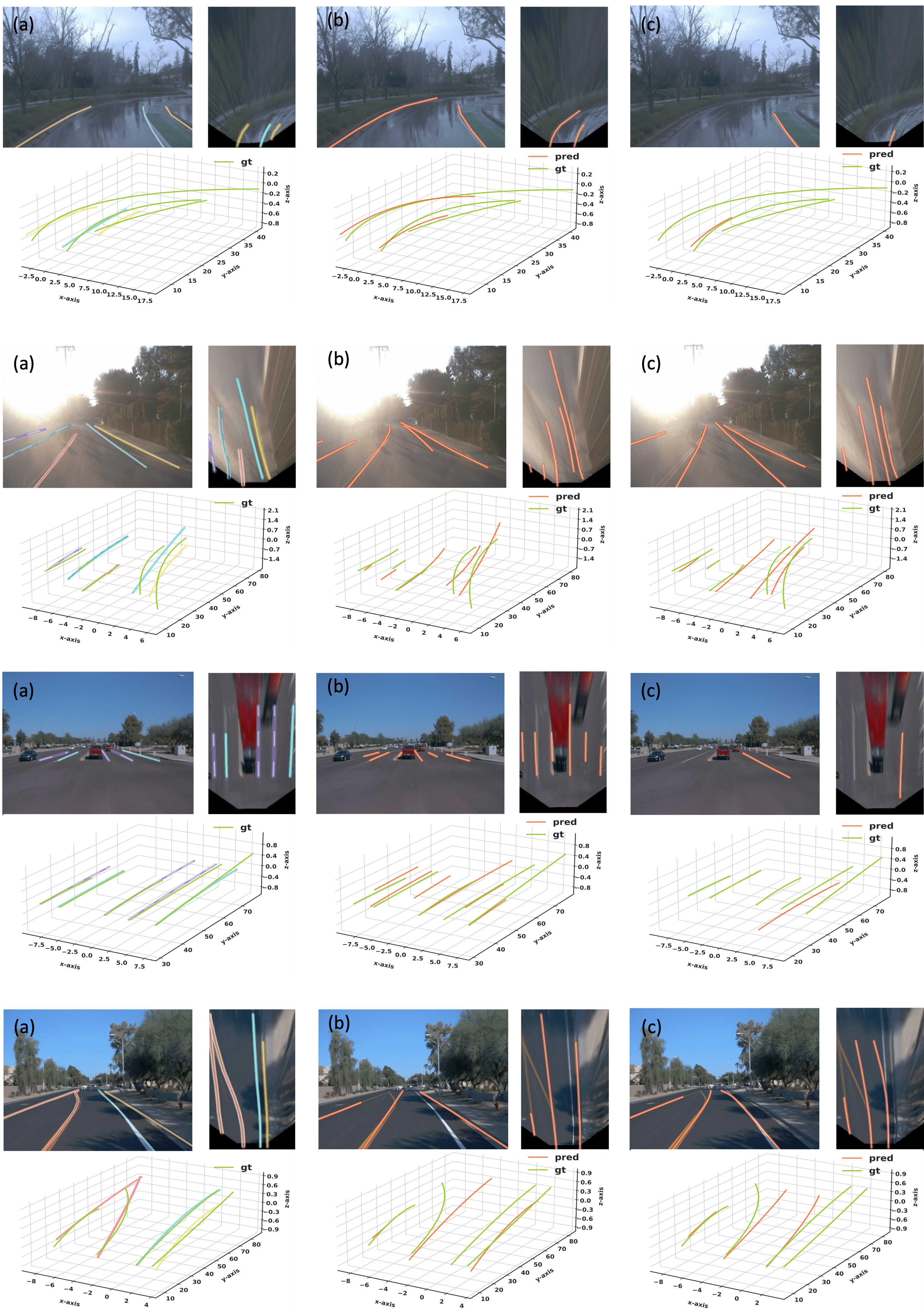}
    \caption{Qualitative results of PersFormer(a), 3D-LaneNet(b) \cite{Garnett_2019_ICCV}, and Gen-LaneNet(c) \cite{guo2020gen} on OpenLane. Extreme weather case, Intersection case and Merge\&Split case}
    \label{fig:qualitative-2}
\end{figure}

\begin{figure}[tb!]
    \centering
    \includegraphics[width=\textwidth]{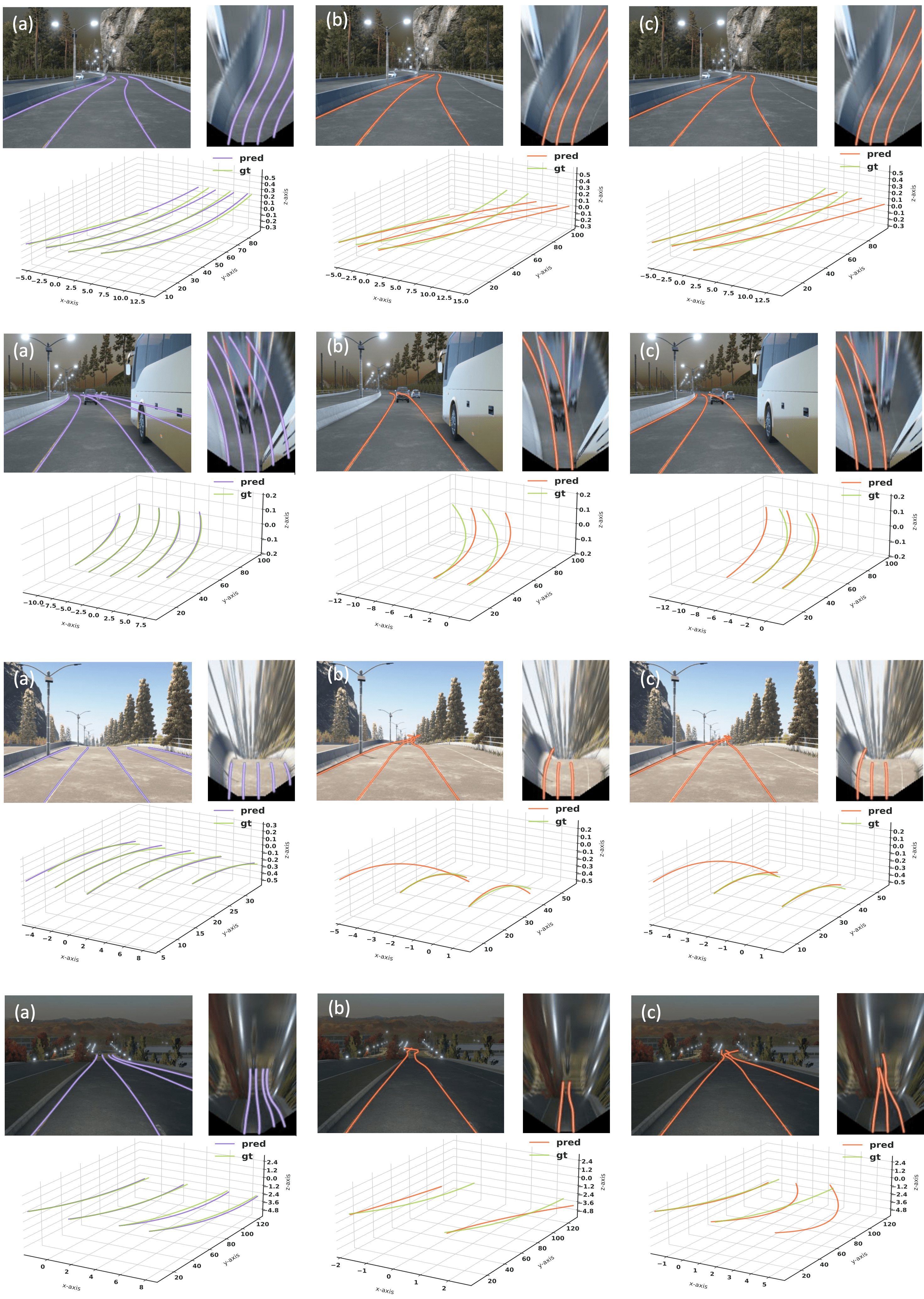}
    \caption{Qualitative results of PersFormer(a), 3D-LaneNet(b) \cite{Garnett_2019_ICCV}, and Gen-LaneNet(c) \cite{guo2020gen} on Apollo. Curve case and Up\&Down case}
    \label{fig:qualitative-apollo}
\end{figure}

We provide qualitative results compared with SOTA 3D lane detection methods in different evaluation scenarios on OpenLane dataset in Fig. \ref{fig:qualitative-1},\ref{fig:qualitative-2}.
Results on Apollo 3D synthetic dataset are shown in Fig. \ref{fig:qualitative-apollo}.
We can observe that PersFormer could achieve higher accuracy and capture more lanes to reconstruct the scenes on both datasets.

\section{License of Assets}
OpenLane dataset is based on the Waymo Open Dataset \cite{sun2020scalability} and therefore we distribute the data under Creative Commons Attribution-NonCommercial-ShareAlike license and Waymo Dataset License Agreement for Non-Commercial Use (August 2019). You are free to share and adapt the data, but have to give appropriate credit and may not use the work for commercial purposes. All code of PersFormer and OpenLane toolkit is under Apache License 2.0.

The pretrained ResNet model weights are under the MIT license. We integrate part of the code of Deformable-DETR \cite{zhu2021deformable} and Gen-LaneNet \cite{guo2020gen} which are under Apache License 2.0. We also use part of the code of LaneATT \cite{tabelini2021keep} which is under the MIT license.

\section{Outlook}

As OpenLane is built upon Waymo Open Dataset \cite{sun2020scalability}, a road-object joint detection framework is possible in the future. 
Moreover, BEV is the necessity in the future of autonomous driving, and how to design a better BEV representation remains to be explored. The proposed PersFormer may also be adapted to new tasks. 

\end{document}